\documentclass[10pt,twocolumn,letterpaper]{article}

\usepackage{wacv}
\usepackage{times}
\usepackage{epsfig}
\usepackage{graphicx}
\usepackage{amsmath}
\usepackage{amssymb}
\usepackage{booktabs}

\usepackage{tikz,pgfplots,multirow}
\usepackage{colortbl}
\usepackage{hhline}
\usepackage[accsupp]{axessibility}

\definecolor{rowgray}{rgb}{0.975,0.975,0.975}
\definecolor{rowlearn}{rgb}{0.9375,0.9375,0.9}
\definecolor{rowexp}{rgb}{0.9175,0.9,0.9}
\definecolor{rowcost}{rgb}{0.9,0.9175,0.9}
\definecolor{rowblue}{rgb}{0.9,0.9,0.9125}
\definecolor{note}{rgb}{0.2, 0.2, 0.8}

\def\wacvPaperID{186} 

\wacvapplicationstrack 

\wacvfinalcopy 

\ifwacvfinal
\usepackage[breaklinks=true,bookmarks=false]{hyperref}
\else
\usepackage[pagebackref=true,breaklinks=true,colorlinks,bookmarks=false]{hyperref}
\fi

\definecolor{blue}{rgb}{0.184313725, 0.333333333, 0.592156863}
\definecolor{linkred}{rgb}{0.517647059, 0.235294118, 0.047058824}
\definecolor{green}{rgb}{0.219607843, 0.341176471, 0.137254902}

 \hypersetup{
 	colorlinks=True,
	linkcolor=linkred,
	filecolor=blue,
	citecolor = green,      
	urlcolor=blue,
}

\begin{document}

\title{Mobile Robot Manipulation using Pure Object Detection}

\author{Brent Griffin\\
Agility Robotics\\
{\tt\small brent.griffin@agilityrobotics.com}
}

\maketitle
\thispagestyle{empty}

\begin{abstract}
   This paper addresses the problem of mobile robot manipulation using object detection. Our approach uses detection and control as complimentary functions that learn from real-world interactions. We develop an end-to-end manipulation method based solely on detection and introduce Task-focused Few-shot Object Detection (TFOD) to learn new objects and settings. Our robot collects its own training data and automatically determines when to retrain detection to improve performance across various subtasks (e.g., grasping). Notably, detection training is low-cost, and our robot learns to manipulate new objects using as few as four clicks of annotation. In physical experiments, our robot learns visual control from a single click of annotation and a novel update formulation, manipulates new objects in clutter and other mobile settings, and achieves state-of-the-art results on an existing visual servo control and depth estimation benchmark. Finally, we develop a TFOD Benchmark to support future object detection research for robotics: \url{https://github.com/griffbr/tfod}.

\end{abstract}

\section{Introduction}
\label{sec:intro}

Object detection, i.e., predicting bounding boxes and category labels for objects in an RGB image, has seen remarkable methodological advances \cite{CaEtAl20,ReEtAl16,ReEtAl15} thanks to an abundance of annotated training and evaluation data in high-quality datasets \cite{EvEtAl15,LVIS,MSCOCO}.
Recently, \textit{few-shot} object detection \cite{ChEtAl18,WaRaHe19,YaEtAl19} has emerged as a critical innovation to detect new object classes using a limited subset of annotated examples from existing datasets \cite{KaEtAl19,WaEtAl20}.

Detection supports many applications \cite{PaPaVi19, XiMa20, ZhEtAl20}.
When changing objects, tasks, or environments, however, we find that off-the-shelf detectors are less reliable outside of their initial training setting, causing the subsequent application to fail.
Thus, passive learning is not enough, especially for robots, where visual experience is dynamic and interactive \cite{BoEtAl17}.
Furthermore, robots that only use passive data are wasting a critical asset---the ability to interact with the world and learn from those interactions.

\begin{figure}
	\centering
	\includegraphics[width=0.45\textwidth]{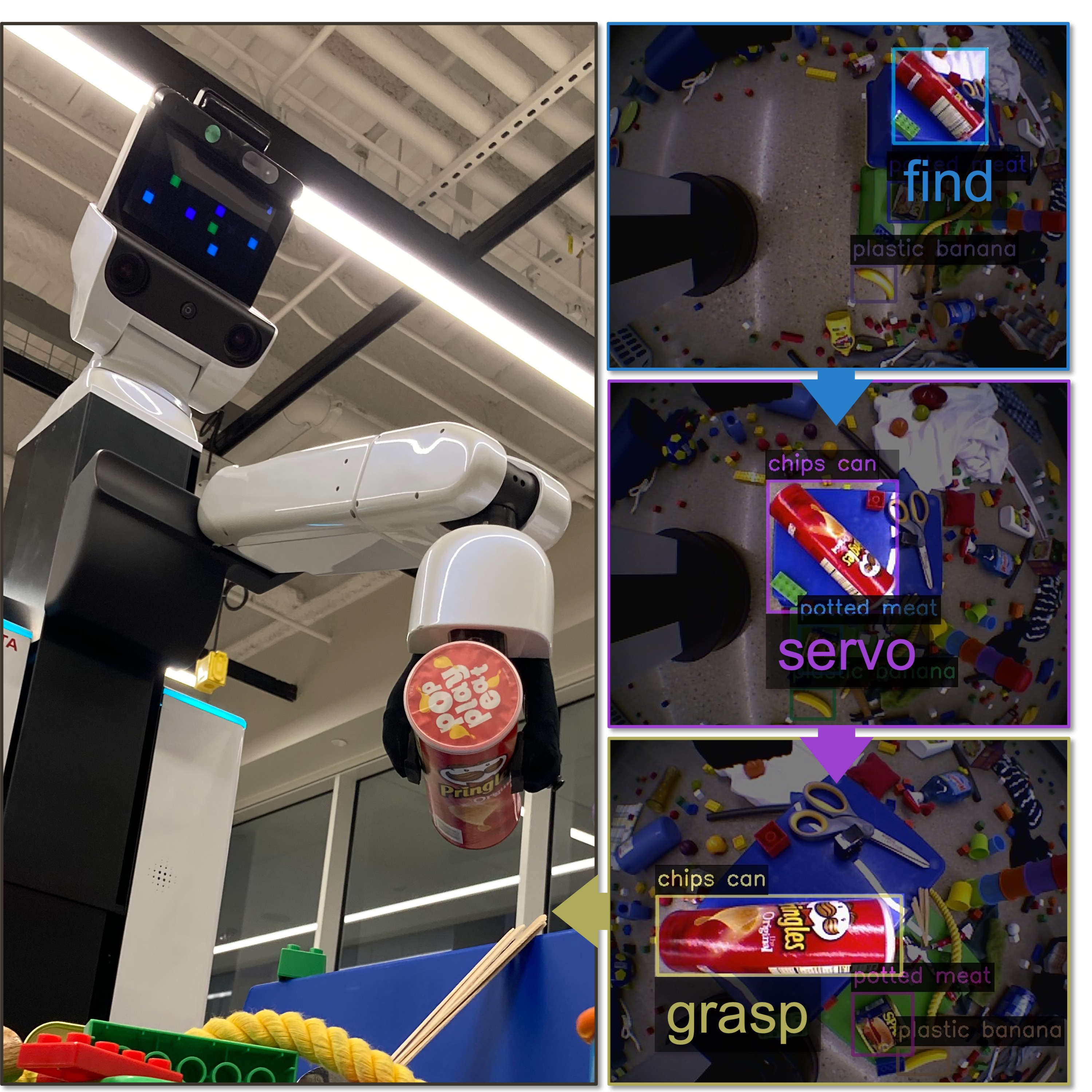}
	\caption{
		\textbf{Detection-based Manipulation}.
		Perceiving objects from camera motion and bounding boxes (right), our robot learns to grasp the Chips Can in four robot-collected training examples.
	}
	\label{fig:front}
\end{figure}

To that end, this paper presents \textit{ClickBot}, a robot that learns mobile manipulation for new objects and changing environments using pure object detection (i.e., learning and perceiving objects using only 2D bounding boxes).
Our approach relies on two primary contributions.

\begin{itemize}
	\item [1)] \textit{Detection-Based Manipulation}. 
	We develop a novel set of detection-based tasks to complete mobile robot manipulation.
	Innovations include a novel update formulation to learn visual servo control, motion-based depth estimation that improves as ClickBot approaches objects, and active multi-view grasp selection.
	Using our approach, ClickBot manipulates unstructured objects without 3D models using a single RGB camera.
	To our knowledge, this is the first work to develop end-to-end object manipulation entirely from detection.
	
	\item [2)] \textit{\textbf{T}ask-Focused \textbf{F}ew-Shot \textbf{O}bject \textbf{D}etection (TFOD)}. 
	We introduce TFOD to learn detection-based tasks for new objects and settings.
	Using TFOD, ClickBot automatically performs tasks, collects data, determines if new few-shot examples are required, and, if so, directs annotation toward specific tasks.
	In practice, TFOD improves performance for difficult or evolving robot tasks while reducing overall annotation costs.
\end{itemize}

We validate our combined approach in a variety of robot experiments. 
First, ClickBot learns detection-based visual control on average in less than 14~\textrm{s} and reduces learning variability by 65-85\% relative to prior visual servo approaches.
Next, ClickBot achieves state-of-the-art results on the VOSVS Benchmark \cite{GrFlCo20}, increasing visual servo control and depth estimation performance by 16.7\% and 25.0\% respectively.
Finally, ClickBot learns to grasp objects in clutter (see Figure~\ref{fig:front}) and cleans up scattered objects with moving placement locations at 124.6 picks-per-hour. 

This paper builds a foundation to guide future research and innovation for using few-shot detection algorithms in robotics.
However, many researchers do not have a robot or data to evaluate their algorithms in a robotics setting.
Thus, as a final contribution, we develop a corresponding TFOD Benchmark. 
The TFOD Benchmark is configurable for a variety of few-shot object detection settings, includes evaluation across a diverse set of YCB Dataset objects \cite{YCB} using standard MS-COCO AP metrics \cite{MSCOCO}, and will guide future research toward increasingly reliable detection in this new task-focused setting for robot manipulation.

\section{Related Work}
\label{sec:related}

\noindent \textbf{Object Detection} is a preliminary process for many methods in our community. Example detection-based methods include segmentation \cite{maskrcnn}, 3D shape prediction \cite{GkMaJo19}, depth \cite{GrCo21} and pose estimation \cite{PaPaVi19, XiMa20}, and single-view metrology \cite{ZhEtAl20}, to name but a few.
In this paper, we similarly introduce a novel detection-based method for mobile robot manipulation that operates directly from object detection.

Learning object detection typically requires a large number of bounding box annotations from a labeled dataset for training and evaluation \cite{EvEtAl15,MSCOCO},
with some datasets additionally focusing on continuous recognition \cite{LoMa17} or multi-view indoor environments \cite{ReMoBa18}.
However, static datasets do not account for new objects and settings in the wild.

\vspace{2mm}
\noindent \textbf{Few-Shot Object Detection} (FSOD) addresses part of this limitation by detecting new objects from only a few annotated examples \cite{ChEtAl18}.
For evaluation, the first FSOD benchmark \cite{KaEtAl19} uses set splits of $k=1, 2, 3, 5, 10$ annotated bounding boxes for 5 few-shot objects on the PASCAL VOC Dataset \cite{EvEtAl15} and $k=10, 30$ for 20 few-shot objects on the MS-COCO Dataset \cite{MSCOCO}.
Subsequent work \cite{WaEtAl20} revises this protocol by randomly selecting few-shot objects and annotated examples with an average evaluation over 40 trials with additional results on the LVIS Dataset \cite{LVIS}.

Using these prior benchmarks, FSOD has seen rampant methodological advances.
Initial finetuning methods treat FSOD as a transfer learning problem from a large source domain to few-shot objects \cite{ChEtAl18, WaEtAl20}.
Other methods use meta-learning algorithms to learn from existing detectors and quickly adapt to few-shot objects, either by using feature rewieghting schemes \cite{KaEtAl19,YaEtAl19} or by using model parameter generation from base classes to efficiently learn few-shot objects \cite{WaRaHe19}.
Other FSOD approaches include a distance metric learning-based classifier \cite{KaEtAl192}, incremental few-shot learning to reduce training requirements \cite{OpPiZi06,PeEtAl20}, \textit{one}-shot detection by matching and aligning target-image-features with query-image-features \cite{OsEtAl20}, plug-and-play detectors to maintain known category performance while learning new concepts \cite{Zhang_2021_WACV}, and an attention-guided cosine margin to mitigate class imbalance \cite{Agarwal_2022_WACV}, to name but a few.

However, generating few-shot examples for new objects or using FSOD to support other applications has drawn scant attention.
One recent work collects new detection training data by teleoperating a UAV \cite{AlSuSu19}, but this approach uses substantially more training examples than current FSOD methods require ($k\gg30$) and does not consider applications other than detection.
On the other hand, one FSOD method \cite{XiMa20} supports viewpoint estimation applications by developing a unified framework that uses arbitrary 3D models of few-shot objects, but this work only detects and estimates viewpoints for objects in existing datasets.

To that end, this paper extends FSOD by improving detection for specific application tasks and collecting new few-shot examples in the wild for new objects and settings using an approach we call \textit{\textbf{T}ask-Focused} \textbf{F}ew-Shot \textbf{O}bject \textbf{D}etection (TFOD).
Furthermore, rather than trying to predict the best set of few-shot examples \textit{a priori}, we let the robot and difficulty of each task decide, thereby limiting annotation to a few relevant examples.
Notably, we use a fine-tuning FSOD method in our TFOD experiments, but, as we will show, TFOD is generalizable across FSOD methods.

\vspace{2mm}
\noindent \textbf{Visual Servo Control} (VS) uses visual data in a servo loop for robot control.
Closed-form VS methods typically relate image features to robot actuators using a \textit{feature Jacobian} \cite{ChHu06,HuHaCo96,WeSaNe87} with advanced methods learning the feature Jacobian directly on the robot \cite{ChHu07,HoAs94,JaFuNe97}.
Closed-form VS can position UAVs \cite{GuHaMa08,McJaCo17} or wheeled robots \cite{LuOrGi08,MaOrPr07} and manipulate objects \cite{JaFuNe97,KiEtAl16,WaLaSi10}.
Although this early VS work demonstrates the utility of VS, these methods rely on structured visual features (e.g., fiducial markers or LED panels).

Subsequent VS methods manipulate non-structured objects using deep learning.
Learning VS manipulation end-to-end can occur entirely on a robot \cite{AbEtAl19,LaRi13,PiGu16} or in simulation with innovative sim-to-real transfer techniques \cite{JaEtAl19,PeMiCh20,ZuEtAl19}.
However, all of these end-to-end methods learn in a fixed workspace and do not address the challenges of \textit{mobile} manipulation, which includes moving cameras, changing environments, and dynamic grasp positioning.

To bridge the gap between learned VS and mobile robot applications, one recent work develops mobile VS with features based on pre-trained video object segmentation \cite{GrFlCo20}. 
However, this approach does not learn new objects, tasks, or environments, which would require a substantial cost to annotate new segmentation masks \cite{JaGr13}.
To that end, this paper develops a novel detection-based approach to mobile VS, which, in comparative experiments, requires less than 5\% the annotation cost while significantly improving performance.
We also develop a new approach to learn VS using the \textit{pseudoinverse} feature Jacobian, which, relative to prior work, learns VS faster and more consistently.
Finally, coupling our mobile VS with TFOD and other tasks lets our robot learn to locate and manipulate new objects efficiently.

\section{Task-Focused Few-Shot Object Detection}
\label{sec:tfod}

We develop an interactive approach we call \textbf{T}ask-Focused \textbf{F}ew-Shot \textbf{O}bject \textbf{D}etection (TFOD) to collect data and learn detection for new objects and applications.

\subsection{Task and Detection Model}
TFOD is generally applicable for any task $T$ and object detection model $D$ that satisfy the following criteria:
\vspace{2mm}
	\newline 1. $T$ observes $n\geq 1$ images $\{I_1, I_2, \dots, I_n\}$.
	\newline 2. $D(I)$ outputs a set of bounding boxes with class labels.
	\newline 3. There are one or more failure criteria $F$ based on $D$.
	\newline 4. If $F$ occurs, $m \geq 1$ images $\{I_{F_1}, I_{F_2}, \dots, I_{F_m}\}$ \newline \indent $\in \{I_1, I_2, \dots, I_n\}$ are saved to log the failure.
	\newline 5. $D$ can update its output predictions given a set of $p \geq 1$ \newline \indent annotated few-shot examples $E(I_{E_1}, I_{E_2}, \dots, I_{E_p})$. \vspace{2mm} \newline 
Using these definitions, the goal of TFOD is to update $D$ until $T$ is completed without $F$ using the minimum $p$.

In plain words, as our robot attempts difficult or evolving detection-based tasks, its detection model can fail.
However, if our robot recognizes a failure ($F$), it is a meaningful opportunity for learning, and our robot saves image data of the failure for annotation ($I_F$).
After we provide annotation ($E$), our robot updates its detection model. 
Notably, criteria for $F$ and selecting $I_F$ can change depending on the specific task, and we provide several examples in Section~\ref{sec:clickbot}.

\subsection{Task-Focused Data Collection}
Given task $T$ and detection model $D$, our robot performs $T$ using $D(I_1), D(I_2), \dots, D(I_n)$.
If $F$ occurs during $T$, our robot selects one or more representative failure images $I_F \in \{I_1, \dots, I_n\}$ for annotation.
For each $I_F$ we annotate, our robot adds it to an aggregate set of annotated few-shot examples $E(I_{E_1}, I_{E_2}, \dots, I_{E_p})$ that update $D$.
In effect, these updates prevent $F$ from recurring and let our robot learn difficult or evolving tasks.
Once $T$ completes without $F$, our robot goes on to complete others task.

In practice, unless objects or settings change, our robot rarely needs an update after $D$ is learned for $T$.
We also find that sharing few-shot examples in $E$ across related tasks reduces the overall number of examples required.

\subsection{Few-Shot Annotation}
\begin{figure}
	\centering
	\includegraphics[width=0.475\textwidth]{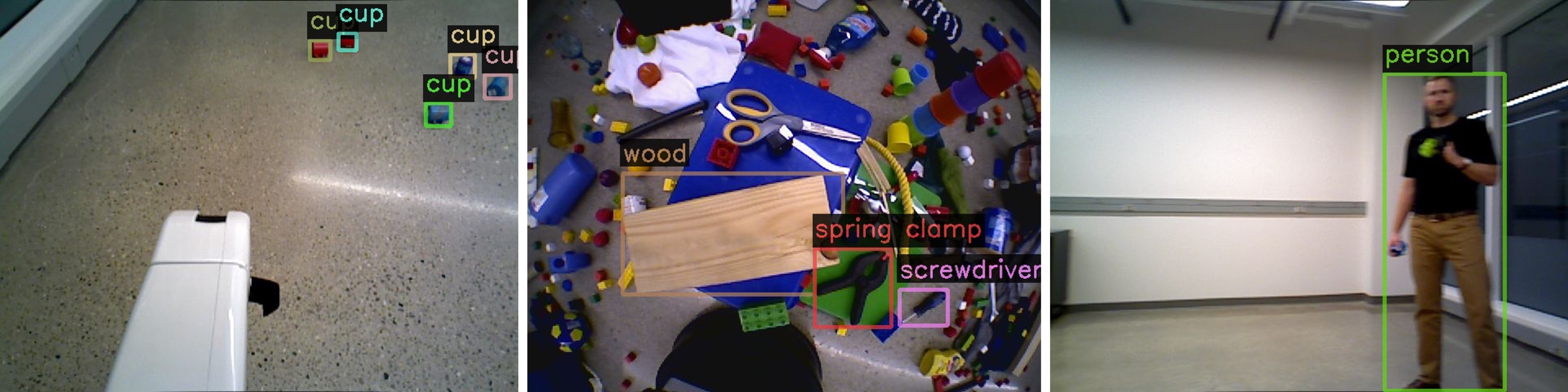} 
	\caption{\textbf{Task-Focused Annotation}.
		ClickBot collects task data and selects new few-shot examples for tasks requiring better detection, e.g., Find (left), Grasp (middle), and Placement (right).
	}
	\label{fig:annotate}
\end{figure}

We provide annotation $E(I_{E_1}, I_{E_2}, \dots, I_{E_p})$ using a custom GUI.
After $F$, a user reviews new failure images $I_F$ saved by our robot.
For each $I_F$ with task-relevant objects, the user can drag bounding boxes around each object and then add the new few-shot example $I_{E}$ to $E$ (see Figure~\ref{fig:annotate}).

In practice, annotating a bounding box takes about 7~\textrm{seconds} per object \cite{JaGr13}.
Also, $I_F$ without objects can \textit{optionally} be added to $E$ as a true negative, which generally reduces false positives from $D$, e.g., for a task that searches for objects.
In this work, we only annotate one $I_F$ with task-relevant objects per update, which gives our robot the opportunity to complete a task using the least annotation.

\section{Detection-Based Manipulation}
\label{sec:clickbot}

We perform mobile robot manipulation using a novel set of detection-based tasks.
Notably, for objects learned \textit{a priori}, our approach also works with standard object detection and tracking algorithms.
To start, our robot needs to find task-relevant objects for manipulation, i.e., the \textit{Find} Task.

\subsection{Finding Objects for Manipulation}

For the Find Task, we use a set of $n$ robot kinematic poses that moves a camera in the task space.
Our robot collects an image at each pose ($I_1, I_2, \dots$) until an object is found using detection model $D$, which completes the Find Task.
Our failure criterion $F$ is if our robot collects all $n$ images without detection, in which case, each image is saved for few-shot annotation ($I_F$).
Using this process, the Find Task is typically how our robot first learns new objects.

In practice, after learning and manipulating new objects, we discontinue the Find Task's failure criterion $F$ to initiate \textit{Sentry Mode}. 
In Sentry Mode, our robot intermittently uses the Find Task to search for objects but no longer assumes that the absence of detections indicates a false negative.
Thus, our robot finds objects if they are in the task space without generating unnecessary few-shot examples if they are absent.

\subsection{Learning Visual Servo Control from Detection}
\label{sec:boxvs}

A key innovation for mobile manipulation is our learned visual servo controller (VS), which enables our robot to position itself relative to found objects, i.e., the \textit{Servo} Task.

\vspace{2mm}
\noindent \textbf{Image Features from Detection}.
To start, we use detection model $D$, input image $I$, and a target object class label $l$ to define image features $\mathbf{s} \in \mathbb{R}^2$ as
\begin{align}
	\mathbf{s}\big(D(I), l, \mathbf{s}_{t-1} \big) := 
	\begin{bmatrix}
		s_x, s_y 
	\end{bmatrix}^\intercal,
	\label{eq:boxfeat}
\end{align}
where bounding boxes with class labels other than $l$ are ignored, $\mathbf{s}_{t-1}$ represents $\mathbf{s}$ from the previous time step, and $s_x, s_y$ denote the two image coordinates of the target object's bounding box center.
We use $\mathbf{s}_{t-1}$ in \eqref{eq:boxfeat} for two reasons.
First, if there are multiple boxes with label $l$, we select the closest match to $\mathbf{s}_{t-1}$ for stability.
Second, we use $\left\| \mathbf{s} - \mathbf{s}_{t-1} \right\|_{L_1}$ to check if $\mathbf{s}$ indicates a physically improbable discontinuity in object position.
Finally, if detection of $l$ is absent at any time step, we temporarily use $\mathbf{s} = \mathbf{s}_{t-1}$.

Using \eqref{eq:boxfeat}, our failure criteria $F$ for the Servo Task are:
\vspace{2mm}
\newline 1. A discontinuity $\left\| \mathbf{s} - \mathbf{s}_{t-1} \right\|_{L_1} > 150$ \textrm{pixels}.
\newline 2. Detection of $l$ is absent for 20 consecutive time steps. \vspace{2mm} \newline
If either of these occur, our robot stops the Servo Task and saves the last input image for few-shot annotation ($I_F$).

\vspace{2mm}
\noindent \textbf{Visual Servo Feedback Control}.
We use image features $\mathbf{s}$ \eqref{eq:boxfeat} for our VS feedback error $\mathbf{e}$, which is defined as
\begin{align}
	\mathbf{e} = \mathbf{s} - \mathbf{s}^* = 
	\begin{bmatrix}
		s_x - s_x^*,~ s_y - s_y^*
	\end{bmatrix}^\intercal,
	\label{eq:e}
\end{align}
where $\mathbf{s}^* \in \mathbb{R}^2$ is the vector of desired feature values.
We also use $\mathbf{s}^*$ to initiate $\mathbf{s}$ at $t=0$ as $\mathbf{s}\big(D(I), l, \mathbf{s}^* \big)$, which starts VS on the target object closest to the desired position.

Typical VS \cite{ChHu06} relates image features $\mathbf{s}$ to six-degrees-of-freedom (6DOF) camera velocity $\mathbf{v}$ using
$\dot{\mathbf{s}} = \mathbf{L_s} \mathbf{v}$,
where $\mathbf{L_s} \in \mathbb{R}^{2 \times 6}$ is called the \textit{feature Jacobian}.
In this work, we use a constant $\mathbf{s}^*$ (i.e., $\dot{\mathbf{s}}^* = 0$), which, from \eqref{eq:e}, implies that $\dot{\mathbf{e}}=\dot{\mathbf{s}}$ and $\dot{\mathbf{s}}= \mathbf{L_s} \mathbf{v}=\dot{\mathbf{e}}$. 
Using this relationship, we find our control input $\mathbf{v}$ to minimize $\mathbf{e}$ using
\begin{align}
	\mathbf{v} = \text{-} \widehat{\mathbf{L}_\mathbf{s}^+} \mathbf{e},
	\label{eq:vs}
\end{align}
where $\widehat{\mathbf{L}_\mathbf{s}^+} \in \mathbb{R}^{6 \times 2}$ is the \textit{pseudoinverse} feature Jacobian. 

When our robot controls $\mathbf{e}$ \eqref{eq:vs} below a threshold, it is accurately positioned relative to an object and the Servo Task is complete. In experiments, we use a threshold of 10 \textrm{pixels} before depth estimation and 5 \textrm{pixels} before grasping.

\vspace{2mm}
\noindent \textbf{A New Update Formulation to Learn Visual Control}.
It is impossible to know the exact feature Jacobian $\mathbf{L_s}$ on real VS systems \cite{ChHu06}.
Instead, some VS work estimates $\mathbf{L_s}$ \cite{HoAs94,JaFuNe97} or $\widehat{\mathbf{L}_\mathbf{s}^+}$ \cite{GrFlCo20} from observations using a Broyden update.
Inspired by \cite[(4.12)]{Br65} in Broyden's original paper, we introduce a new update formulation to estimate $\widehat{\mathbf{L}_\mathbf{s}^+}$ in \eqref{eq:vs} as
\begin{align}
	\widehat{\mathbf{L}_\mathbf{s}^+}_{t+1} := \widehat{\mathbf{L}_\mathbf{s}^+}_t + \alpha \bigg( 
	\frac{ \big(\Delta \mathbf{x} - \widehat{\mathbf{L}_\mathbf{s}^+}_t \Delta \mathbf{e} \big) \Delta  \mathbf{e}^\intercal}
	{  \Delta  \mathbf{e}^\intercal \Delta \mathbf{e} } \bigg) \circ \mathbf{H},
	\label{eq:hb}
\end{align}
where $\alpha \in \mathbb{R}$ determines the update speed, $\Delta \mathbf{x} = \mathbf{x}_t -  \mathbf{x}_{t-1}$ is the change in 6DOF camera position since the last update, $\Delta  \mathbf{e} = \mathbf{e}_t -  \mathbf{e}_{t-1}$ is the change in error, and the element-wise product with logical matrix $\mathbf{H} \in \mathbb{R}^{6 \times 2}$ determines which $\widehat{\mathbf{L}_\mathbf{s}^+}$ elements can update.
We add $\mathbf{H}$ to prevent association of unrelated elements in $\mathbf{v}$ and $\mathbf{e}$ \eqref{eq:vs}, which leads to more consistent and faster learning in comparative experiments.

Our robot actively learns to relate detection to camera motion for visual control \eqref{eq:vs} using \eqref{eq:hb}.
In plain words, our robot moves a camera ($\Delta \mathbf{x}$), observes corresponding changes in detection-based error ($\Delta \mathbf{e}$), then updates its learned motion-detection model ($\widehat{\mathbf{L}_\mathbf{s}^+}$) based on the difference between the actual ($\Delta \mathbf{x}$) and predicted ($\widehat{\mathbf{L}_\mathbf{s}^+}_t \Delta \mathbf{e}$) change in camera position.

In our experiments, we initiate \eqref{eq:hb} with $\widehat{\mathbf{L}_\mathbf{s}^+}_{t=0} = 0_{6 \times 2}$, \newline $\alpha=0.5$, and $\mathbf{H} = \begin{bmatrix}
	0 & 1 & 0 & 0 & 0 & 0 \\ 1 & 0 & 0 & 0 & 0 & 0
\end{bmatrix}^\intercal$.
This choice of $\mathbf{H}$ couples image features $s_x$ and $s_y$ in $\mathbf{e}$ \eqref{eq:e} with the $x$- and $y$-axis camera velocities in $\mathbf{v}$ \eqref{eq:vs}.

\subsection{Estimating Depth from Motion and Detection}
\label{sec:depth}

After centering the camera on an object using the Servo Task, our robot estimates the object's depth using active perception with detection, i.e., the \textit{Depth} Task.

Recent work estimates the depth of detected objects by comparing changes in camera pose to changes in bounding box size \cite[(9)]{GrCo21}.
In this work, we improve this estimate by actively advancing the robot's camera toward an object while recalculating the object's depth using every available detection $D(I_1), \dots, D(I_n)$ with its corresponding kinematic camera pose.
Once our robot estimates the object is within 0.2~\textrm{m}, the Depth Task is complete.
Our failure criterion $F$ for the Depth Task is if detection of the object's label $l$ disappears, in which case, our robot stops the Depth Task and saves the last input image for annotation ($I_F$).

For depth-based grasping, our robot uses the median of an aggregate set of depth estimates, which consists of the latest estimate at every 0.05~\textrm{m} of camera motion.
Basically, this approach mitigates any proximity-based detection errors that can occur when the camera is close to an object.

\subsection{Grasping with Active Perception and Detection}
\label{sec:grasp}

After estimating an object's depth, our robot grasps the object using detection, i.e., the \textit{Grasp} Task.
Similar to other work, we use a simple visual representation that generalizes grasping across many novel objects \cite{VoPrVe19} but also use multiple views to improve grasp selection \cite{MoCoLe20}.

For active grasp planning, our robot moves its camera 0.16~\textrm{m} above the object's estimated depth then uses VS to center the object underneath its gripper. 
Next, our robot rotates its camera to find the best fit between the object and detection bounding boxes.
Bounding boxes are rectangular, so our robot only rotates the camera $\frac{\pi}{2}$ \textrm{radians} because 1) the height at any angle $\theta$ is the same as the width at $\theta + \frac{\pi}{2}$ and 2) the box dimensions at $\theta$ and $\theta + \pi$ are the same.
As in the Depth Task, our Grasp Task failure criterion $F$ is if detection of the object disappears, which causes our robot to stop and save the last input image for annotation ($I_F$).
After rotation and detection, our robot uses the box with the overall minimum height or width to plan its grasp.

\begin{figure}
	\centering
	\includegraphics[width=0.48\textwidth]{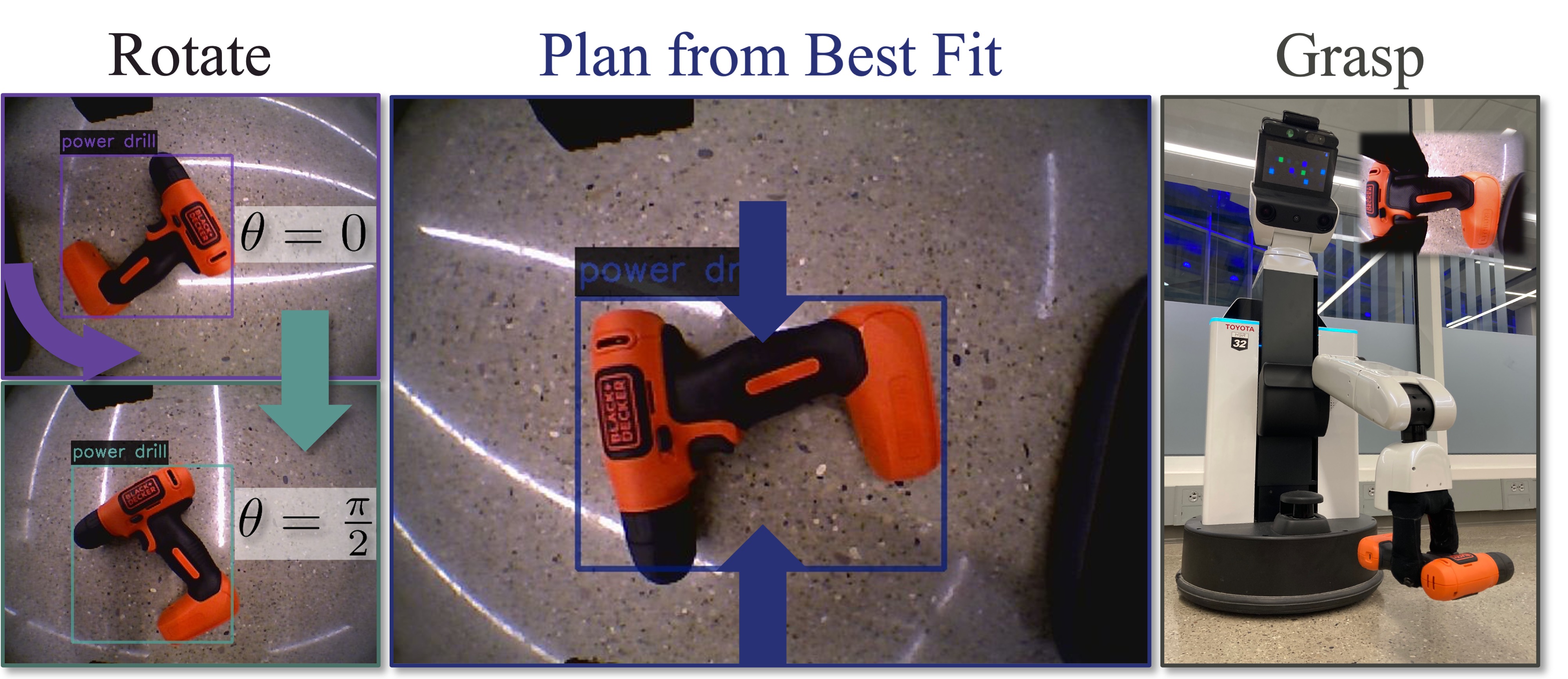}
	\caption{\textbf{Grasping from Detection}.
		ClickBot rotates its camera (left) to find the narrowest detection-based parallel grasp points (middle) then uses a force-based grasp and lifts the Drill (right).
	}
	\label{fig:grasp}
\end{figure}

Our grasp plan uses an antipodal grasp (i.e., a parallel grasp closing on two points).
Basically, our robot uses the narrowest set of detection-based parallel grasp points and grasps at the object's center for balance (see Figure~\ref{fig:grasp}).
After rotating its open gripper to align with the minimum height or width, our robot lowers its gripper to the object's estimated depth and applies a force-based parallel grasp.
Our robot then lifts the object while continuing to apply force.
If the gripper fingers remain separated by the object, the grasp is a success, and our robot releases the object at a goal location before returning to the Find Task for other objects.

\section{Experimental Results}
\label{sec:experiments}

We validate TFOD and detection-based manipulation (ClickBot) in a variety of robot experiments with videos available at \url{https://youtu.be/Bby4Unw7HrI}. 

\subsection{Experimental Setup}
\label{sec:exp_setup}

\noindent \textbf{Robot and Camera Hardware}.
We use a Toyota Human Support Robot (HSR) for our experiments \cite{HSR_journal}.
We detect objects using HSR's end effector-mounted wide-angle grasp camera, which streams 640$\times$480 RGB images at 25~\textrm{Hz}.
We grasp detected objects using HSR's end effector-mounted parallel gripper with series elastic fingertips, which have a 135~\textrm{mm} maximum width.
HSR's end effector moves on a 4DOF manipulator arm mounted on a torso with prismatic and revolute joints, but the relative pose between the grasp camera and gripper are constant.
We typically point the end effector at the ground for detection and grasping (see Figures~\ref{fig:front} and \ref{fig:grasp}).
For mobility, HSR uses a differential drive base.
HSR's base also has a torso revolute joint directly above it, so we can control HSR as an omnidirectional robot (i.e., 3DOF ground-plane translation and rotation).
We use quadratic programming \cite{ShBuHu15} to command camera velocities $\mathbf{v}$ \eqref{eq:vs}, but any velocity controller is applicable.

\vspace{2mm}
\noindent \textbf{Detection Model}.
\label{sec:detection}
For our baseline model, we use Faster R-CNN \cite{ReEtAl15}, which runs in real time and has improved since its original publication.
For reproducibility, we use the same Faster R-CNN configuration as Detectron2 \cite{detectron2} with ResNet~50 pre-trained on ImageNet and a FPN backbone trained on MS-COCO \cite{detectron2}.
In our experiments, we update our detection model using annotated few-shot examples $E$ (Section~\ref{sec:tfod}), which consists of fine-tuning from the baseline model for 1,000 training iterations and takes less than four minutes using a standard workstation and GPU (GTX 1080 Ti).
We also use a relatively high 0.9 confidence score threshold for detection, which significantly decreases false positives at the cost of increasing false negatives.

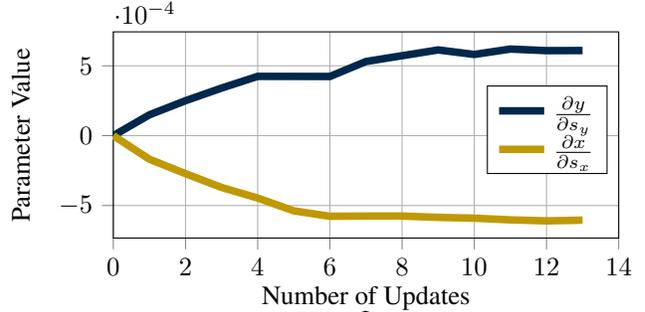
\begin{figure}
	\centering
\begin{tikzpicture}

\definecolor{color0}{rgb}{0.12156862745098,0.466666666666667,0.705882352941177}
\definecolor{color0}{rgb}{0.55,0.3,0.3}
\definecolor{color1}{rgb}{0.00000,0.15290,0.29800} 
\definecolor{color1}{rgb}{0.3,0.3,0.3} 

\definecolor{color0}{rgb}{0.75,0.6,0.015} 
\definecolor{color1}{rgb}{0.00000,0.15290,0.29800} 


\begin{axis}[
width=8.3cm,
height=4.325cm,
tick align=outside,
tick pos=left,
x grid style={lightgray!92.026143790849673!black},
xlabel={Number of Updates},
xmajorgrids,
xmin=0, xmax=14,
y grid style={lightgray!92.026143790849673!black},
ylabel={Parameter Value},
ymajorgrids,
grid style={lightgray!50},
legend style={at={(0.978,0.74)},anchor=north east},
legend cell align=left,
/tikz/inner sep to outer sep/.style={inner sep=0pt, outer sep=.3333em},
x tick label style=inner sep to outer sep,
x label style=inner sep to outer sep,
y label style=inner sep to outer sep,
]

\addplot [line width = 1mm, color1]
table [row sep=\\]{%
	0  0  \\
	1	0.00014908  \\
	2	0.00025015 \\
	3	0.00034058 \\
	4	0.00042442 \\
	5	0.00042398 \\
	6	0.00042357 \\
	7	0.00053076  \\
	8	0.00057221  \\
	9	0.00061358  \\
	10  0.00058128 \\
	11	0.00061995 \\
	12	0.00060857 \\
	13	0.00060927 \\
};
\addlegendentry{$\frac{\partial y}{\partial s_y}$};


\addplot [line width = 1mm, color0]
table [row sep=\\]{%
	0	0  \\
	1	-0.00016851 \\
	2	-0.00027179 \\
	3	-0.00037079 \\
	4	-0.00044657 \\
	5	-0.00053809 \\
	6	-0.00057704\\
	7	-0.00057551 \\
	8	-0.00057536 \\
	9	-0.00058485 \\
	10	-0.0005906 \\
	11	-0.0006027  \\
	12  -0.00061003 \\
	13 -0.00060461	\\
};
\addlegendentry{$\frac{\partial x}{\partial s_x}$};

\end{axis}

\end{tikzpicture}
	\caption{\textbf{Learned Visual Control $\widehat{\mathbf{L}_\mathbf{s}^+}$ Parameter Convergence}. 
	}
	\label{fig:learn_param}
\end{figure}

\subsection{Learning Visual Servo Control from One Click}
\label{sec:exp_learn}

ClickBot learns visual servo control (VS) from camera motion and detection using our new update formulation \eqref{eq:hb}.
For each VS learning experiment, ClickBot starts a motion sequence, tracks detection changes, and updates $\widehat{\mathbf{L}_\mathbf{s}^+}$ after each motion.
Each learning experiments ends when $\widehat{\mathbf{L}_\mathbf{s}^+}$ converges, i.e., $\left\| \widehat{\mathbf{L}_\mathbf{s}^+}_{t+1} - \widehat{\mathbf{L}_\mathbf{s}^+}_{t} \right\|_{L_1} < 10^{-6}$.

For camera motion ($\Delta \mathbf{x}$), ClickBot repeats eight motions comprising the permutations of \{-5, 0, 5\} \textrm{cm} across the $x$ and $y$ axes (e.g., $x=$-5, $y=$5). 
These motions are varied yet cycle through the initial camera pose for continued learning.

For detection, we use the racquetball from the YCB Object Dataset \cite{YCB}.
The racquetball is placed underneath ClickBot's grasp camera and our detection model learns from a single bounding box (i.e., one click of annotation).
Notably, ClickBot learns VS from detection error \textit{changes} $\Delta \mathbf{e}$ \eqref{eq:hb}, so the \textit{constant} desired values $s^*$ in $\mathbf{e}$ \eqref{eq:e} are arbitrary.

In addition to learning VS for our remaining ClickBot experiments, we perform a single consecutive set of trials to compare \eqref{eq:hb} against existing $\widehat{\mathbf{L}_\mathbf{s}^+}$ update formulations.
Notably, two formulas use $\Delta \mathbf{x}^\intercal \widehat{\mathbf{L}_\mathbf{s}^+}_t \Delta \mathbf{e}$ in the denominator and are undefined for $\widehat{\mathbf{L}_\mathbf{s}^+}_{t=0} = 0_{6 \times 2}$.
Thus, for VOSVS \cite[(11)]{GrFlCo20} and Broyden \cite[(4.5)]{Br65}, we use VOSVS's convention and initiate with $\widehat{\mathbf{L}_\mathbf{s}^+}_{t=0} = \begin{bmatrix}
	0 & 1 & 0 & 0 & 0 & 0 \\ 1 & 0 & 0 & 0 & 0 & 0
\end{bmatrix}^\intercal \cdot 10^{-3}$.

\vspace{2mm}
\noindent \textbf{Results}.
ClickBot learns the VS model that we use in our remaining experiments in 13.29~\textrm{s} with 13 Broyden updates.
Immediately afterward, we push the racquetball and ClickBot follows it, confirming that the learned visual controller \eqref{eq:vs} is a success (see video in supplementary material).
We plot the learned $\widehat{\mathbf{L}_\mathbf{s}^+}$ values at each update in Figure~\ref{fig:learn_param}.

We provide comparative VS learning results in Table~\ref{tab:vs_compare}.
Relative to prior formulations, ClickBot requires 30-60\% the updates and has 15-35\% as much overall learned parameter variation.
Thus, our new update formulation \eqref{eq:hb} learns VS faster and more reliably than the prior formulations.

\setlength{\tabcolsep}{1.4pt} 
\begin{table} [t]
	\centering
	\caption{
		\textbf{Visual Servo Learning Results}
		are from a single consecutive set of 10 trials for each update formulation.
	}
	\footnotesize
	\begin{tabular}{| l | c | r | c | r | r | c | }
		\hline
		\multicolumn{1}{| c |}{\cellcolor{rowexp}} & \multicolumn{2}{c |}{\cellcolor{rowlearn}Updates} & \multicolumn{4}{c|}{\cellcolor{rowcost}Range of $\widehat{\mathbf{L}_\mathbf{s}^+}$ Parameter}\\
		\multicolumn{1}{| c |}{\cellcolor{rowexp}$\widehat{\mathbf{L}_\mathbf{s}^+}$ Update} & \multicolumn{2}{c |}{\cellcolor{rowlearn} Required} & \multicolumn{4}{c|}{\cellcolor{rowcost}Values Learned ($\cdot 10^{-4}$)}\\ 
		\hhline{| >{\arrayrulecolor{rowexp}}- >{\arrayrulecolor{black}}| ------}
		\multicolumn{1}{| c |}{\cellcolor{rowexp}Equation} & \multicolumn{1}{c |}{\cellcolor{rowlearn}Mean} & \multicolumn{1}{c|}{\cellcolor{rowlearn}Range} & \cellcolor{rowcost}$\frac{\partial x}{\partial s_x}$ & \multicolumn{1}{c|}{\cellcolor{rowcost}$\frac{\partial x}{\partial s_y}$} & \multicolumn{1}{c|}{\cellcolor{rowcost}$\frac{\partial y}{\partial s_x}$} & \multicolumn{1}{c|}{\cellcolor{rowcost}$\frac{\partial y}{\partial s_y}$} \\ \hline
		ClickBot \eqref{eq:hb}	&	\bf 13.6	&	\bf 9--21~	&	-6.2--	-5.6	&	\bf 0.0--0.0	&	\bf 0.0--0.0	&	\bf 5.4--6.1	\\
		\rowcolor{rowgray} VOSVS \cite[(11)]{GrFlCo20} &	22.5	&	15--30~	&	\bf -6.1-- -5.6	&	\bf 0.0--0.0	&	\bf 0.0--0.0	&	5.4--8.3 \\
		Broyden \cite[(4.12)]{Br65}	&	31.7	&	15--76~	&	-6.4--	-5.8	&	-0.2--0.3	&	-1.0--1.5	&	5.4--7.2	\\
		\rowcolor{rowgray} Broyden \cite[(4.5)]{Br65}	& 45.8	&	21--101	&	-6.5--	-5.6	&	-0.4--0.8	&	-1.1--2.2	&	5.5--8.3	\\
		\hline
	\end{tabular}
	\label{tab:vs_compare}
\end{table}

\begin{figure}
	\centering
	\includegraphics[width=0.475\textwidth]{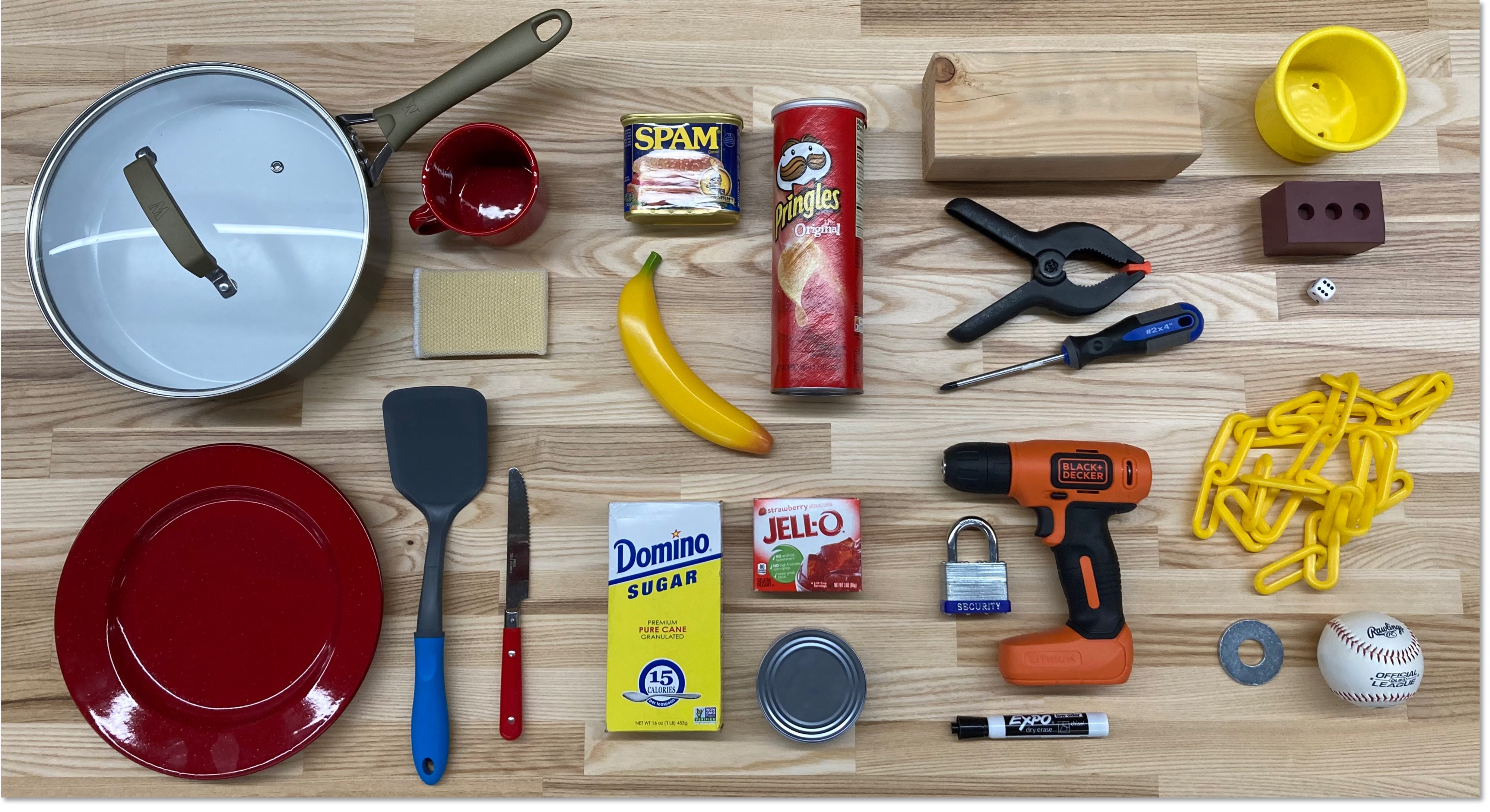}
	\caption{\textbf{Experiment Objects from YCB Dataset}.
		Object sets left to right are Kitchen, Food, Tool, and Shape.
		Dimensions span between 4--470~\textrm{mm} and many objects exhibit specular reflection.
	}
	\label{fig:ycb}
\end{figure}

\subsection{VOSVS Benchmark}
\label{sec:exp_trial}

We evaluate ClickBot's VS and active depth estimation 
using the VOSVS Benchmark \cite{GrFlCo20}.
This benchmark consists of eight consecutive trials of VS and depth estimation (DE) on the YCB objects \cite{YCB} shown in Figure~\ref{fig:ycb}.
Each trial starts with three objects supported at 0.0, 0.125, and 0.25~\textrm{m} above the ground within camera view.
VS is a success if a robot locates and servos to an object for DE.
DE is a success if a robot advances without collision then closes it's gripper on the object without hitting the underlying surface.

\setlength{\tabcolsep}{5.6pt}
\begin{table}
	\centering
	\caption{\textbf{VOSVS Benchmark Results}
		use a single consecutive set of trials. 
		Visual Servo (VS) is a success (\checkmark) if a robot moves to an object for depth estimation (DE). DE is a success if a robot's gripper closes on an object without collision.
	}
	\footnotesize
	\begin{tabular}{ | l | l | l | c | c | c | c |}
		\hline 
		\multicolumn{1}{|c|}{\cellcolor{rowblue}} & \multicolumn{1}{c|}{\cellcolor{rowblue}Support} & \multicolumn{1}{c|}{\cellcolor{rowblue}} & \multicolumn{4}{c|}{\cellcolor{rowexp}Method} \\
		\hhline{| >{\arrayrulecolor{rowblue}}- >{\arrayrulecolor{black}}| >{\arrayrulecolor{rowblue}}- >{\arrayrulecolor{black}}|>{\arrayrulecolor{rowblue}}- >{\arrayrulecolor{black}}| ----|}
		\multicolumn{1}{|c|}{\cellcolor{rowblue}Object} & \multicolumn{1}{c|}{\cellcolor{rowblue}Height} & \multicolumn{1}{c|}{\cellcolor{rowblue}YCB} & \multicolumn{2}{c|}{\cellcolor{rowexp}ClickBot} & \multicolumn{2}{c|}{\scriptsize \cellcolor{rowexp}VOSVS \cite{GrFlCo20}} \\
		\hhline{| >{\arrayrulecolor{rowblue}}- >{\arrayrulecolor{black}}| >{\arrayrulecolor{rowblue}}- >{\arrayrulecolor{black}}|>{\arrayrulecolor{rowblue}}- >{\arrayrulecolor{black}}| ----|}
		\multicolumn{1}{|c|}{\cellcolor{rowblue}Set} & \multicolumn{1}{c|}{ \cellcolor{rowblue}(\textrm{m})} & \multicolumn{1}{c|}{\cellcolor{rowblue}Object \cite{YCB}} & \cellcolor{rowexp}VS & \cellcolor{rowexp}DE & \cellcolor{rowexp}VS & \cellcolor{rowexp}DE \\
		\hline								
		Tool	&	0.25 &Power Drill &	\checkmark & 	&	\checkmark	&	\checkmark 	\\
		\rowcolor{rowgray} 	Tool	&	0.125	&	Marker &	\checkmark & 	&\checkmark &		\\
		Tool	&	0.0	&	Padlock&	\checkmark & \checkmark	&\checkmark	&	 	\\  \hline			
		\rowcolor{rowgray} 	Tool	&	0.25	&	Wood&	\checkmark & \checkmark	&\checkmark	&	 	\\
		Tool	&	0.125	&	Spring Clamp&	\checkmark & 	&\checkmark	&	 	\\
		\rowcolor{rowgray} 	Tool	&	0.0	& Screwdriver&	\checkmark & \checkmark	&	\checkmark	&	 	\\  \hline
		Food	&	0.25	&Chips Can&	\checkmark & \checkmark	&\checkmark	&	\checkmark 	\\
		\rowcolor{rowgray} 	Food	&	0.125	&	Potted Meat&	\checkmark & \checkmark	&\checkmark	&	\checkmark 	\\
		Food	&	0.0	&	Plastic Banana&	\checkmark & \checkmark	&\checkmark	&	\checkmark 	\\					 \hline		
		\rowcolor{rowgray} Food	&	0.25	&	Box of Sugar&	\checkmark & \checkmark	&	\checkmark	&	\checkmark 	\\
		Food	&	0.125	&	Tuna&	\checkmark & \checkmark	&\checkmark	&	 	\\
		\rowcolor{rowgray} 	Food	&	0.0	&	Gelatin&	\checkmark & \checkmark	&\checkmark	&	\checkmark 	\\	 \hline		
		Kitchen	&	0.25	&	Mug&	\checkmark & \checkmark	&\checkmark	&	\checkmark 	\\
		\rowcolor{rowgray} 	Kitchen 	&	0.125	& Softscrub&	\checkmark &	&		&	 	\\
		Kitchen	&	0.0	&	Skillet with Lid&	\checkmark & 	&	&	 	\\	 \hline						
		\rowcolor{rowgray} 	Kitchen	&	0.25	&Plate&	\checkmark & \checkmark	&	\checkmark	&	\checkmark 	\\
		Kitchen	&	0.125	&	Spatula	&	\checkmark &&	&	  	\\
		\rowcolor{rowgray} 	Kitchen	&	0.0	&	Knife&	\checkmark & \checkmark	&\checkmark	&	 	\\	 \hline			
		Shape	&	0.25	&Baseball&	\checkmark & \checkmark	&	\checkmark	&	 	\\
		\rowcolor{rowgray} 	Shape	&	0.125	&	Plastic Chain &	\checkmark & \checkmark	&\checkmark	&		\\
		Shape	&	0.0	&	Washer&	\checkmark & 	&\checkmark	&	 	\\		 \hline						
		\rowcolor{rowgray} 	Shape	&	0.25	&	Stacking Cup&	\checkmark & \checkmark	&\checkmark	&	\checkmark 	\\
		Shape	&	0.125	&	Dice&	\checkmark & 	&	&	 	\\
		\rowcolor{rowgray} 	Shape	&	0.0	&Foam Brick&	\checkmark & \checkmark	&	\checkmark	&	\checkmark 	\\
		\hline
		\hline	\multicolumn{3}{| c |}{\cellcolor{rowcost}Success Rate (\%)} & \bf 100 & \bf 66.7& 83.3 & 41.7  \\ \hline 
		\hline
		\rowcolor{rowgray} \multicolumn{3}{| c |}{\cellcolor{rowlearn}Annotations Per Object} & \multicolumn{2}{c|}{\bf 3.7} & \multicolumn{2}{c|}{10} \\ \hline
		\multicolumn{3}{| c |}{\cellcolor{rowlearn}Annotation Time Per Object} & \multicolumn{2}{c|}{\bf 26 \textrm{s}} & \multicolumn{2}{c|}{540 \textrm{s}} \\ \hline
	\end{tabular}
	\label{tab:benchmark}
\end{table}

\setlength{\tabcolsep}{9pt} 
\begin{table*} [t]
	\centering
	\caption{
		\textbf{Task-Focused Few-Shot Annotation Results} are averaged across corresponding trials (individual results in supplementary material).
		Clicks are the number of annotated bounding boxes, which each require 7~\textrm{s} (see user study \cite{JaGr13}).
		CPU refers to training time.
	}
	\footnotesize
	\begin{tabular}{| l | c | c | c | c | c | c | r | r | r |}
		\hline
		\cellcolor{rowblue}  & \multicolumn{5}{c |}{\cellcolor{rowlearn}Number of Task-Focused} & \multicolumn{4}{c |}{\cellcolor{rowcost}Requirements Per Object Class} \\
		\hhline{| >{\arrayrulecolor{rowblue}}- >{\arrayrulecolor{black}}| >{\arrayrulecolor{rowlearn}} *{5}{-}>{\arrayrulecolor{black}}*{4}{|-}}
		\cellcolor{rowblue} & \multicolumn{5}{ c |}{\cellcolor{rowlearn}Few-Shot Examples Generated ($E$)} & \multicolumn{2}{c |}{\cellcolor{rowcost}Annotation}  & \multicolumn{1}{c |}{\cellcolor{rowcost}Robot}& \multicolumn{1}{c |}{\cellcolor{rowcost}CPU} \\
		\hhline{| >{\arrayrulecolor{rowblue}}- >{\arrayrulecolor{black}}| ---------}
		\multicolumn{1}{| c |}{\cellcolor{rowblue}Task-Focused Learning Experiment} & \cellcolor{rowlearn}Find & \cellcolor{rowlearn}Servo &  \cellcolor{rowlearn}Depth & \cellcolor{rowlearn}Grasp & \cellcolor{rowlearn}Total & \multicolumn{1}{c|}{\cellcolor{rowcost}Clicks} &  \multicolumn{3}{c |}{\cellcolor{rowcost}Time (\textrm{seconds})}  \\
		\hline 
		Learning Visual Servo Control  & 1.0 & 0.0 & N/A & N/A & 1.0 & 1.0 & 7.0   & 13.3 & 227  \\ 
		\rowcolor{rowgray} 
		VOSVS Benchmark & 1.0	&	0.9	&	3.1	&	N/A	&	5.0	 &	3.7	&	26.0	&	20.2	&	 383  	  \\ 
		Pick-and-Place with Prior Annotation 	& 0.3	&	0.3	&	1.3	&	2.8	&	4.5	&	3.4	&	23.9	&	29.1	&	343  \\ 
		\rowcolor{rowgray} Pick-and-Place \textit{in Clutter} with Prior Annotation & 	0.5	&	0.8	&	0.0	&	2.3	&	3.5	&	2.7	&	18.7	&	23.2	&	287  \\ 
		Pick-and-Place  & 1.0	&	0.8	&	2.5	&	3.8	&	8.0	&	6.0	&	42.0	&	51.4	&	615  \\ 
		\rowcolor{rowgray} Pick-and-Place \textit{in Clutter} & 1.0	&	2.0	&	4.3	&	3.3	&	10.5	&	7.5	&	52.5	&	67.3	&	811  \\ 
		\hline
	\end{tabular}
	\label{tab:learn}
\end{table*}

We also use the VOSVS Benchmark to evaluate TFOD-based learning.
ClickBot learns new objects for each trial using the Find, Servo, and Depth tasks from Section~\ref{sec:clickbot}. 
Starting without any annotation, ClickBot's first few-shot example $E(I_{E_1})$ is from an initial Find pose, and ClickBot returns to the Find Task after any other vision updates.

For each trial object, ClickBot finds it (Find), servos to it until $\mathbf{e} < 10$ \textrm{pixels} \eqref{eq:vs} (Servo), descends within an estimated 0.2~\textrm{m} (Depth), then closes it's gripper at the estimated depth.
Each object is removed after its first full attempt, i.e., Find, Servo, Depth, and grasp closure without an update.

\vspace{2mm}
\noindent \textbf{Results}.
We provide comparative VOSVS Benchmark results in Table~\ref{tab:benchmark}.
ClickBot achieves a perfect VS score and improves the prior DE success rate from 42\% to 67\%.
ClickBot is perfect on the Food set but leaves room to improve DE on the Tool and Kitchen sets by 50\%.

We also compare annotation time in Table~\ref{tab:benchmark}.
A segmentation mask takes about 54~\textrm{s} to annotate \cite{JaGr13}, which equates to VOSVS using 540~\textrm{s} of annotation per object.
On the other hand, ClickBot uses a simpler bounding box-based representation with task-focused annotation, which equates to 26~\textrm{s} of annotation per object, a 95\% reduction.

We provide detailed TFOD results in Table~\ref{tab:learn}.
ClickBot averages 5 updates per trial with more few-shot examples for Depth than Find and Servo combined.
A primary goal of TFOD is to focus annotation on difficult tasks, so we are encouraged that ClickBot automatically identifies and directs annotation to the task that requires the most improvement.

\subsection{Pick-and-Place in Cluttered Environments}
\label{sec:exp_grasp}

We evaluate ClickBot's end-to-end manipulation by adding pick-and-place to the VOSVS Benchmark for a new set of consecutive trials.
First, we add the full Grasp Task (Section~\ref{sec:grasp}) after VS and DE for the Tool and Food sets.
Notably, HSR cannot physically grasp some Kitchen and Shape objects, e.g., because they are too heavy (Skillet with Lid) or too low to the ground (Washer).
After grasping, ClickBot also attempts to place objects in a bin.
For evaluation, Grasp is only considered a success if ClickBot moves the object without dropping it and then releases it in the bin.
Finally, as an added challenge, we repeat all of the consecutive pick-and-place trials in a cluttered environment. 

We also use these pick-and-place trials to test two ablative TFOD configurations.
For the first ablative configuration, we modify ClickBot to start with prior annotation from Section~\ref{sec:exp_trial} for the non-cluttered pick-and-place trials.
Subsequently, any new annotation is also included when learning pick-and-place in clutter.
For a second ablative configuration, we remove TFOD and ClickBot \textit{only} uses the prior annotation.
For this configuration, ClickBot also uses a 0.1 confidence score threshold to increase detection likelihood.

\setlength{\tabcolsep}{6.3pt}
\begin{table}
	\centering
	\caption{\textbf{Pick-and-Place Results} with the Tool and Food Sets from the VOSVS Benchmark.
		All results use a single RGB camera. 
	}
	\footnotesize
	\begin{tabular}{ | l | c | c | c | c | c |}
		\hline 
		\cellcolor{rowexp}   & \multicolumn{2}{ c |}{\cellcolor{rowlearn}Annotation}& \multicolumn{3}{ c |}{\cellcolor{rowcost}Success Rate (\%)} \\
		\hhline{| >{\arrayrulecolor{rowexp}} - >{\arrayrulecolor{black}} | ----- |}
		\multicolumn{1}{|c|}{\cellcolor{rowexp}Method}  & \cellcolor{rowlearn}Prior & \cellcolor{rowlearn}TFOD & \cellcolor{rowcost}VS & \cellcolor{rowcost}DE & \cellcolor{rowcost}Grasp \\
		\hline \multicolumn{6}{|c|}{\scriptsize \cellcolor{rowblue}VOSVS Benchmark for Tool and Food Sets} \\ \hline							
		VOSVS \cite{GrFlCo20} & Yes & No & \bf 	100 & 50 & N/A	\\ 
		\rowcolor{rowgray} ClickBot & No & Yes & \bf 100 & \bf 75 & N/A \\ \hline
		\multicolumn{6}{|c|}{\scriptsize \cellcolor{rowblue}Tool and Food Sets \textit{with} Pick-and-Place Added} \\ \hline
		ClickBot without TFOD  & Yes & No & 92 & 75 & 50 \\ 
		\rowcolor{rowgray} ClickBot with Prior & Yes & Yes & \bf 100 & \bf 100 & \bf 75 \\ 
		ClickBot & No & Yes & \bf 100 & \bf 100 & \bf 75 \\ \hline
		\multicolumn{6}{|c|}{\scriptsize \cellcolor{rowblue}Tool and Food Sets \textit{with} Pick-and-Place in Clutter} \\ \hline	
		ClickBot without TFOD & Yes & No & 75	&	67	&	58 \\ 
		\rowcolor{rowgray} ClickBot with Prior & Yes & Yes & \bf 100	&	\bf 100	& 69 \\ 
		ClickBot & No & Yes & \bf 100	& \bf 100	&	\bf 88 \\ \hline
	\end{tabular}
	\label{tab:grasp}
\end{table}

\begin{figure*}
	\centering
	\includegraphics[width=1\textwidth]{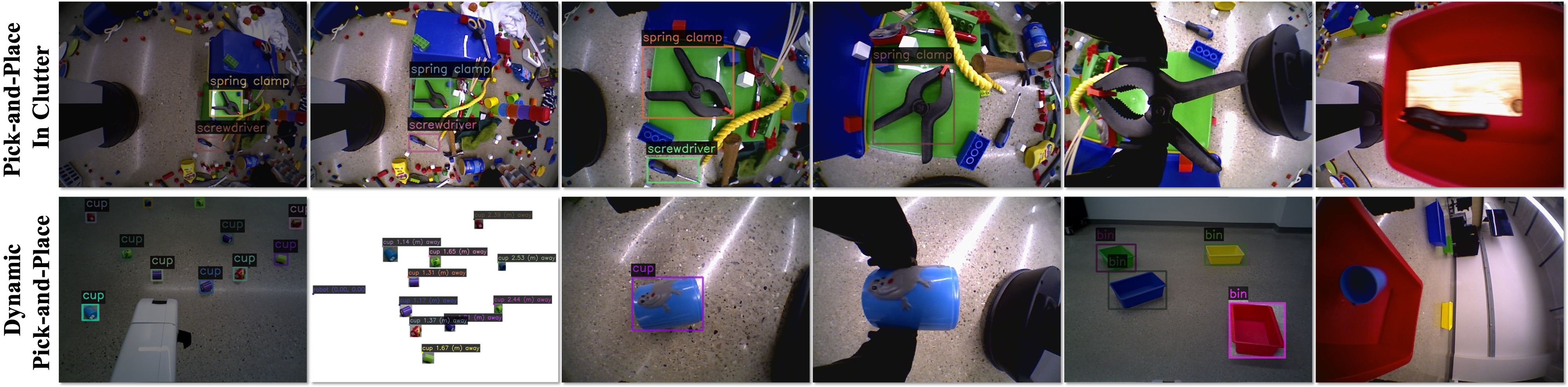}
	\caption{\textbf{Experimental Results}. 
		For pick-and-place in clutter (top), ClickBot uses motion and detection to estimate the spring clamp's depth (left) and active detection-based grasping to place it in a bin (right).
		In dynamic pick-and-place (bottom), ClickBot uses detection with its head camera to map and grasp scattered objects (left) and then similarly discovers a suitable placement location (right).
	}
	\label{fig:exp}
\end{figure*}

\vspace{2mm}
\noindent \textbf{Results}.
We provide ablative pick-and-place results in Table~\ref{tab:grasp}.
The standard configuration achieves the best cluttered Grasp and pick-and-place rate of 88\% (see two results in Figures~\ref{fig:front} and \ref{fig:exp}).
Considering the learning results in Table~\ref{tab:learn}, we attribute the performance difference of the standard configuration over its ablative counterparts to having the most few-shot examples in clutter, which improves task performance for that particular setting.
Nonetheless, the standard configuration uses less than a minute of annotation per object, which is approximately the same amount of time required to annotate a single segmentation mask and much less than the time required to generate a 3D model.

Across all tasks and setting in Table~\ref{tab:grasp}, using TFOD improves performance.
Both ClickBot configurations using TFOD were perfect for VS and DE regardless of clutter.
As in Section~\ref{sec:exp_trial}, ClickBot primarily requests annotation for tasks that require improvement, particularly when using prior annotation, which focuses most new annotation on grasping.
Notably, Grasp-focused annotation can also improve detection performance in other tasks, such as DE.


\subsection{Pick-and-Place with Dynamic Locations}
\label{sec:exp_pph}

We perform qualitative experiments to evaluate ClickBot's mobile manipulation with dynamic placement locations.
For dynamic placement, ClickBot performs a second Find Task using a new set of placement class labels (e.g., Bin or Person).
Once a placement location is detected, ClickBot releases the grasped object at that location.

We also use these experiments to demonstrate the modularity of ClickBot tasks.
Using detection with HSR's RGBD head camera, ClickBot now creates a map for grasp and placement objects during the Find Task.
This map effectively replaces the Depth Task while all other tasks remain.

\vspace{2mm}
\noindent \textbf{Results}.
For our first dynamic pick-and-place experiment, we scatter cups for grasping and bins for placement across the floor.
ClickBot learns to grasp a cup after two few-shot examples and learns to place it in a bin after two more (we show this result in Figure~\ref{fig:exp}, bottom).
We attribute four-shot dynamic pick-and-place to removing the Depth Task, which offsets annotation on the new placement-based Find Task.

For our second experiment, ClickBot learns to retrieve thrown cups and return them to a moving person using eight more few-shot examples (see Figure~\ref{fig:annotate} right).

As a final demonstration, ClickBot places scattered cups in specific bins that match colors.
As an added challenge, we move the bins after each placement.
ClickBot places all nine cups in their correct bin at a rate of 124.6 picks-per-hour.
To our knowledge, there is no precedent for this rate of vision-based mobile robot manipulation in the literature.

\setlength{\tabcolsep}{6.575pt}
\begin{table}
	\centering
	\caption{\textbf{Task-Focused Few-Shot Object Detection Benchmark}
		evaluation uses MS-COCO AP metrics and $k$ few-shot examples.
	}
	\footnotesize
	\begin{tabular}{ | c | c | c | c | c | c | c | c |}
		\hline 
		\cellcolor{rowexp}Method  & \cellcolor{rowlearn}$k$  & \cellcolor{rowcost}AP & \cellcolor{rowcost}AP50 & \cellcolor{rowcost}AP75 & \cellcolor{rowcost}APs & \cellcolor{rowcost}APm &  \cellcolor{rowcost}APl \\
		\hline
		& 1 & 14.1 &19.9
		&17.2
		&0.0
		&32.9 &	22.8 \\
		\rowcolor{rowgray} ClickBot & 2 & 18.3
		& 24.3
		&	22.5
		& 0.0
		& 32.1 &
		27.7 \\
		& 4 & 35.0
		& 46.0 & 42.0 & 1.7 & 57.4 & 39.0 \\ \hline
	\end{tabular}
	\label{tab:tfod}
\end{table}

\subsection{Task-Focused Few-Shot Detection Benchmark}
\label{sec:tfod_bench}

ClickBot's performance will improve with future few-shot object detection methods.
Thus, we are introducing the \textbf{T}ask-Focused \textbf{F}ew-Shot \textbf{O}bject \textbf{D}etection (TFOD) Benchmark to help guide innovation.
The TFOD Benchmark is configurable for $k=1,2,4$ annotated bounding boxes across 12 YCB \cite{YCB} object classes, and our test set includes challenging examples in cluttered settings.
The TFOD Benchmark makes robot-collected data and corresponding annotations publicly available for research, which enables object detection researchers to evaluate their methods in this new task-focused setting for robot manipulation.

\vspace{2mm}
\noindent \textbf{Results}.
We provide baseline TFOD results in Table~\ref{tab:tfod}, which averages our fine-tuning approach (Section~\ref{sec:detection}) across ten consecutive trials (per-object baseline results in supplementary material).
We see opportunity for future object detection innovation across all settings, especially for small objects (APs) and one- or two-shot detection.

\section{Conclusions}
We develop a new method of mobile manipulation based on object detection.
To our knowledge, our robot is the first to manipulate objects using detection alone.
Furthermore, our robot collects data as it performs tasks and, if it recognizes a detection error, automatically selects a new few-shot example for annotation to improve performance.
In this way, our robot avoids many vision-based errors while adapting to changing objects, tasks, and environments.

We evaluate our approach using a variety of experiments.
First, our robot learns a novel visual servo controller from detection in 13.3~\textrm{s}.
Furthermore, we show in repeat trials that our visual servo formulation learns faster and more reliably than alternative approaches.
Using learned visual control with detection-based depth estimation, our robot also achieves state-of-the-art results on an existing visual servo control and depth estimation benchmark.
Next, our robot learns to grasp objects in clutter using a single RGB camera with as few as four few-shot examples, achieving an overall pick-and-place rate of 88\%.
This result is on par or better than recent state-of-the-art methods \cite{MoCoLe20,PeMiCh20,VoPrVe19,ZeEtAl20}, which all use an RGBD camera in a fixed workspace. 
Notably, we can optionally configure our approach for an RGBD input, which our robot uses to clean up scattered objects with moving placement locations at over 120 picks-per-hour.

In conclusion, our experiments show that our RGB-based approach to mobile manipulation works if few-shot annotation is acceptable to learn new objects and settings.
In addition, our approach can supplement RGBD-based approaches or substitute when full 3D sensing is unavailable.

In future work, we will expand our approach to accommodate new challenging tasks (e.g., manipulation across multiple cluttered rooms). 
Future innovations in object detection will help us achieve these results.
Thus, we are releasing a new object detection benchmark that enables future detection work to evaluate and improve performance in a challenging robotics setting.
We also plan to release future additions for this benchmark in new application areas.

{\small
\bibliographystyle{ieee_fullname}
\bibliography{cvpr_click_bot}
}


\clearpage

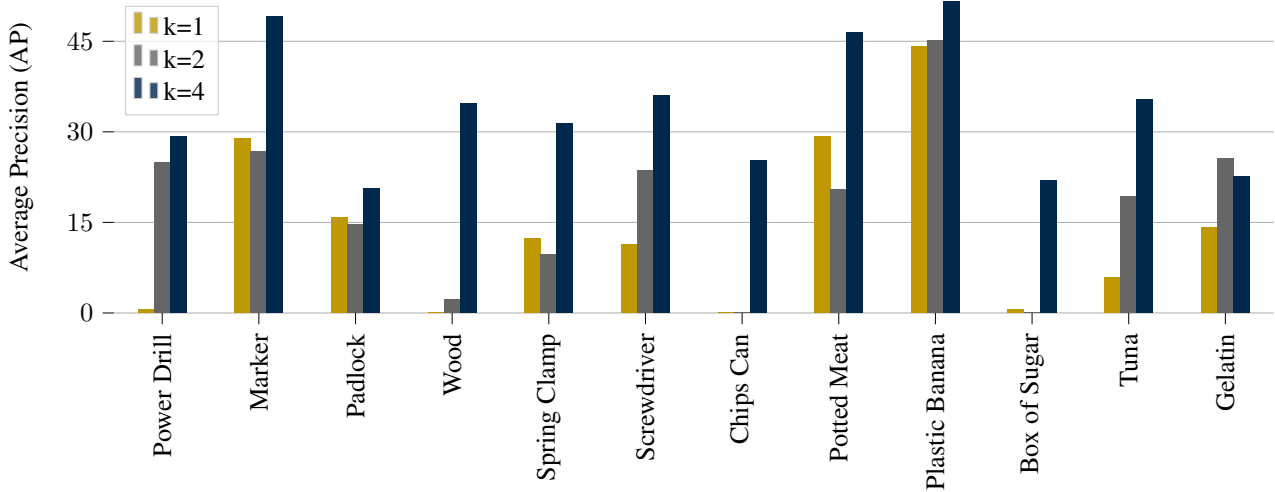
\begin{figure*}[h!]
	\centering
\begin{tikzpicture}

\definecolor{color0}{rgb}{0.12156862745098,0.466666666666667,0.705882352941177}
\definecolor{color1}{rgb}{1,0.498039215686275,0.0549019607843137}
\definecolor{color2}{rgb}{0.172549019607843,0.627450980392157,0.172549019607843}

\definecolor{color0}{rgb}{0.55,0.3,0.3}
\definecolor{color1}{rgb}{0.3,0.55,0.3}
\definecolor{color2}{rgb}{0.3,0.3,0.55}

\definecolor{color0}{rgb}{0.6,0.3,0.3}
\definecolor{color1}{rgb}{0.4,0.4,0.4}
\definecolor{color2}{rgb}{0.3,0.3,0.6}

\definecolor{color0}{rgb}{0.75,0.6,0.015} 
\definecolor{color2}{rgb}{0.00000,0.15290,0.29800} 

\begin{axis}[
	width=17cm,
	height=6cm,
ylabel={Average Precision (AP)},
legend cell align={left},
legend style={
  fill opacity=0.8,
  draw opacity=1,
  text opacity=1,
  at={(0.05,0.925)},
  anchor=north,
  draw=white!80!black
},
ymajorgrids,
tick align=outside,
tick pos=left,
x grid style={white!69.0196078431373!black},
xmin=-0.5, xmax=11.5,
axis line style={draw=none},
xtick style={color=black},
xtick={0,1,2,3,4,5,6,7,8,9,10,11},
ytick={0,15,30,45},
xticklabel style={rotate=90.0},
xticklabels={
  Power Drill,
  Marker,
  Padlock,
  Wood,
  Spring Clamp,
  Screwdriver,
  Chips Can,
  Potted Meat,
  Plastic Banana,
  Box of Sugar,
  Tuna,
  Gelatin
},
y grid style={white!69.0196078431373!black},
ymin=0, ymax=55,
ytick style={color=black}
]
\draw[draw=none,fill=color0] (axis cs:-0.25,0) rectangle (axis cs:-0.0833333333333333,0.526);
\addlegendimage{ybar,ybar legend,draw=none,fill=color0}
\addlegendentry{k=1}

\draw[draw=none,fill=color0] (axis cs:0.75,0) rectangle (axis cs:0.916666666666667,28.886);
\draw[draw=none,fill=color0] (axis cs:1.75,0) rectangle (axis cs:1.91666666666667,15.882);
\draw[draw=none,fill=color0] (axis cs:2.75,0) rectangle (axis cs:2.91666666666667,0);
\draw[draw=none,fill=color0] (axis cs:3.75,0) rectangle (axis cs:3.91666666666667,12.32);
\draw[draw=none,fill=color0] (axis cs:4.75,0) rectangle (axis cs:4.91666666666667,11.273);
\draw[draw=none,fill=color0] (axis cs:5.75,0) rectangle (axis cs:5.91666666666667,0);
\draw[draw=none,fill=color0] (axis cs:6.75,0) rectangle (axis cs:6.91666666666667,29.233);
\draw[draw=none,fill=color0] (axis cs:7.75,0) rectangle (axis cs:7.91666666666667,44.1);
\draw[draw=none,fill=color0] (axis cs:8.75,0) rectangle (axis cs:8.91666666666667,0.616);
\draw[draw=none,fill=color0] (axis cs:9.75,0) rectangle (axis cs:9.91666666666667,5.941);
\draw[draw=none,fill=color0] (axis cs:10.75,0) rectangle (axis cs:10.9166666666667,14.129);
\draw[draw=none,fill=color1] (axis cs:-0.0833333333333333,0) rectangle (axis cs:0.0833333333333333,24.917);
\addlegendimage{ybar,ybar legend,draw=none,fill=color1}
\addlegendentry{k=2}

\draw[draw=none,fill=color1] (axis cs:0.916666666666667,0) rectangle (axis cs:1.08333333333333,26.816);
\draw[draw=none,fill=color1] (axis cs:1.91666666666667,0) rectangle (axis cs:2.08333333333333,14.671);
\draw[draw=none,fill=color1] (axis cs:2.91666666666667,0) rectangle (axis cs:3.08333333333333,2.252);
\draw[draw=none,fill=color1] (axis cs:3.91666666666667,0) rectangle (axis cs:4.08333333333333,9.609);
\draw[draw=none,fill=color1] (axis cs:4.91666666666667,0) rectangle (axis cs:5.08333333333333,23.532);
\draw[draw=none,fill=color1] (axis cs:5.91666666666667,0) rectangle (axis cs:6.08333333333333,0);
\draw[draw=none,fill=color1] (axis cs:6.91666666666667,0) rectangle (axis cs:7.08333333333333,20.443);
\draw[draw=none,fill=color1] (axis cs:7.91666666666667,0) rectangle (axis cs:8.08333333333333,45.182);
\draw[draw=none,fill=color1] (axis cs:8.91666666666667,0) rectangle (axis cs:9.08333333333333,0);
\draw[draw=none,fill=color1] (axis cs:9.91666666666667,0) rectangle (axis cs:10.0833333333333,19.247);
\draw[draw=none,fill=color1] (axis cs:10.9166666666667,0) rectangle (axis cs:11.0833333333333,25.515);
\draw[draw=none,fill=color2] (axis cs:0.0833333333333333,0) rectangle (axis cs:0.25,29.174);
\addlegendimage{ybar,ybar legend,draw=none,fill=color2}
\addlegendentry{k=4}

\draw[draw=none,fill=color2] (axis cs:1.08333333333333,0) rectangle (axis cs:1.25,49.078);
\draw[draw=none,fill=color2] (axis cs:2.08333333333333,0) rectangle (axis cs:2.25,20.673);
\draw[draw=none,fill=color2] (axis cs:3.08333333333333,0) rectangle (axis cs:3.25,34.659);
\draw[draw=none,fill=color2] (axis cs:4.08333333333333,0) rectangle (axis cs:4.25,31.336);
\draw[draw=none,fill=color2] (axis cs:5.08333333333333,0) rectangle (axis cs:5.25,36.082);
\draw[draw=none,fill=color2] (axis cs:6.08333333333333,0) rectangle (axis cs:6.25,25.253);
\draw[draw=none,fill=color2] (axis cs:7.08333333333333,0) rectangle (axis cs:7.25,46.379);
\draw[draw=none,fill=color2] (axis cs:8.08333333333333,0) rectangle (axis cs:8.25,51.535);
\draw[draw=none,fill=color2] (axis cs:9.08333333333333,0) rectangle (axis cs:9.25,21.906);
\draw[draw=none,fill=color2] (axis cs:10.0833333333333,0) rectangle (axis cs:10.25,35.412);
\draw[draw=none,fill=color2] (axis cs:11.0833333333333,0) rectangle (axis cs:11.25,22.656);
\end{axis}

\end{tikzpicture}
	\caption{\textbf{Per-Object Task-Focused Few-Shot Object Detection (TFOD) Benchmark Results}. All TFOD test results correspond to the baseline ClickBot method in Table~\ref{tab:tfod}. There are many opportunities for future improvements, especially for $k=1,2$ few-shot examples.
	}
	\label{fig:TFOD_bar}
\end{figure*}

\begin{figure}
	\centering
\begin{tikzpicture}

\definecolor{color0}{rgb}{0.12156862745098,0.466666666666667,0.705882352941177}
\definecolor{color0}{rgb}{0.55,0.3,0.3}
\definecolor{color1}{rgb}{0.00000,0.15290,0.29800} 
\definecolor{color1}{rgb}{0.3,0.3,0.3} 

\definecolor{color0}{rgb}{0.75,0.6,0.015} 
\definecolor{color2}{rgb}{0.00000,0.15290,0.29800} 


\begin{axis}[
width=8.3cm,
height=6cm,
tick align=outside,
tick pos=left,
x grid style={lightgray!92.026143790849673!black},
xmajorgrids,
xmin=0, xmax=14,
y grid style={lightgray!92.026143790849673!black},
ylabel={$\Delta \mathbf{x}$ (\textrm{m})},
ymajorgrids,
xticklabels={},
grid style={lightgray!50},
ymin=-0.06, ymax=0.04,
legend style={at={(0.97,0.02)},anchor=south east},
legend cell align=left,
/tikz/inner sep to outer sep/.style={inner sep=-5pt, outer sep=0em},
ytick={-0.06,-0.04,-0.02,0,0.02,0.04},
y label style=inner sep to outer sep,
]

\addplot [line width = 0.5mm, lightgray!92.026143790849673!black, forget plot]
table [row sep=\\]{
0 0 \\
14 0 \\
};

\addplot [line width = 1.5mm, draw opacity=0.65, color0]
table [row sep=\\]{%
0	0  \\
1	0.02636285	\\
2	-0.02709418	\\
3	-0.02368997	\\
4	0.02749131	\\
5	-0.02776984	\\
6	0.02711017	\\
7	-9.374E-05	\\
8	-0.00082157	\\
9	0.02664894	\\
10	-0.02685497	\\
11	-0.02504967	\\
12	0.02756886	\\
13	-0.02666232	\\
};
\addlegendentry{$\text{ClickBot Base Forward}$};

\addplot [line width = 1.5mm, draw opacity=0.65, color1]
table [row sep=\\]{%
0	0	\\
1	0.02786655	\\
2	0.02541102	\\
3	-0.02261121	\\
4	-0.02738099	\\
5	-0.00126952	\\
6	-4.775E-05	\\
7	-0.02556659	\\
8	0.02638745	\\
9	0.02697975	\\
10	0.02404704	\\
11	-0.02261136	\\
12	-0.02653534	\\
13	-0.00263152	\\
};
\addlegendentry{$\text{ClickBot Base Lateral}$};

\end{axis}

\end{tikzpicture}
\begin{tikzpicture}

\definecolor{color0}{rgb}{0.12156862745098,0.466666666666667,0.705882352941177}
\definecolor{color0}{rgb}{0.55,0.3,0.3}
\definecolor{color1}{rgb}{0.00000,0.15290,0.29800} 
\definecolor{color1}{rgb}{0.3,0.3,0.3} 

\definecolor{color0}{rgb}{0.75,0.6,0.015} 
\definecolor{color1}{rgb}{0.00000,0.15290,0.29800} 


\begin{axis}[
width=8.3cm,
height=4.325cm,
tick align=outside,
tick pos=left,
x grid style={lightgray!92.026143790849673!black},
xlabel={Number of Updates},
xmajorgrids,
xmin=0, xmax=14,
y grid style={lightgray!92.026143790849673!black},
ylabel={Parameter Value},
ymajorgrids,
grid style={lightgray!50},
legend style={at={(0.978,0.74)},anchor=north east},
legend cell align=left,
/tikz/inner sep to outer sep/.style={inner sep=0pt, outer sep=.3333em},
x tick label style=inner sep to outer sep,
x label style=inner sep to outer sep,
y label style=inner sep to outer sep,
]

\addplot [line width = 1mm, color1]
table [row sep=\\]{%
	0  0  \\
	1	0.00014908  \\
	2	0.00025015 \\
	3	0.00034058 \\
	4	0.00042442 \\
	5	0.00042398 \\
	6	0.00042357 \\
	7	0.00053076  \\
	8	0.00057221  \\
	9	0.00061358  \\
	10  0.00058128 \\
	11	0.00061995 \\
	12	0.00060857 \\
	13	0.00060927 \\
};
\addlegendentry{$\frac{\partial y}{\partial s_y}$};


\addplot [line width = 1mm, color0]
table [row sep=\\]{%
	0	0  \\
	1	-0.00016851 \\
	2	-0.00027179 \\
	3	-0.00037079 \\
	4	-0.00044657 \\
	5	-0.00053809 \\
	6	-0.00057704\\
	7	-0.00057551 \\
	8	-0.00057536 \\
	9	-0.00058485 \\
	10	-0.0005906 \\
	11	-0.0006027  \\
	12  -0.00061003 \\
	13 -0.00060461	\\
};
\addlegendentry{$\frac{\partial x}{\partial s_x}$};

\end{axis}

\end{tikzpicture}
	\caption{
		\textbf{Learned Visual Control $\widehat{\mathbf{L}_\mathbf{s}^+}$ Parameter Convergence with Camera Movement}. 
		ClickBot learns detection-based visual servo control in 13.3~\textrm{seconds} after 13 camera movements (top) and corresponding Broyden updates \eqref{eq:hb} (bottom).
		Subsequently, ClickBot uses this learned visual control in all other experiments.
	}
	\label{fig:learn_vs_cam}
\end{figure}
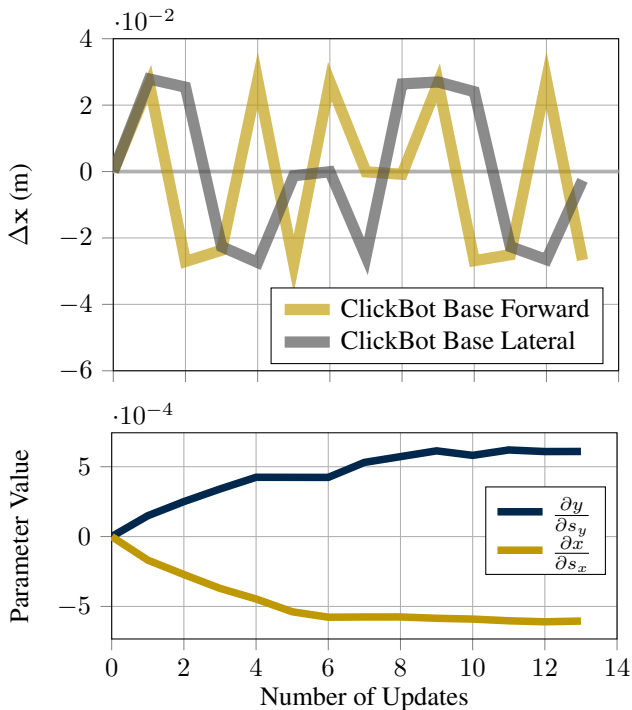

\begin{figure}
	\centering
\begin{tikzpicture}

\definecolor{color0}{rgb}{0.12156862745098,0.466666666666667,0.705882352941177}
\definecolor{color1}{rgb}{0.75,0.35,0.35} 
\definecolor{color0}{rgb}{0.00000,0.15290,0.29800} 

\begin{axis}[
	width=8cm,
	height=8cm,
ylabel={Estimated Object Depth (\textrm{m})},
xlabel={Camera Movement Distance (\textrm{m})},
tick align=outside,
tick pos=left,
x grid style={white!69.0196078431373!black},
xmin=0, xmax=0.15,
xtick style={color=black},
y grid style={white!69.0196078431373!black},
ymin=0.15, ymax=0.3,
xmajorgrids,
ymajorgrids,
legend style={at={(0.9875,0.98)},anchor=north east},
legend cell align=left,
xtick={0,0.05,0.1,0.15},
xticklabels={0,0.05,0.1,0.15},
ytick={0.15,0.2,0.25, 0.3},
yticklabels={0.15,0.2,0.25, 0.3},
ytick style={color=black}
]
\addplot [only marks, mark=*, draw=color0, fill=color0, colormap/viridis, mark size=2pt, draw opacity=0.225, fill opacity=0.55]
table{%
x  y
0.0235790610313416 0.942120468622268
0.0276549458503723 0.810569149319443
0.0315101742744446 0.613801962528616
0.033277302980423 0.589536357084381
0.0363926887512207 0.496208373827513
0.0380370020866394 0.470216028858214
0.0405210554599762 0.429971854196238
0.0418371558189392 0.393388656770618
0.0429742634296417 0.349762617410148
0.0445614457130432 0.325885789784008
0.0455755293369293 0.314466062686289
0.0468942821025848 0.314688329845565
0.0476002097129822 0.294792284143769
0.0483778715133667 0.291634115233447
0.048786073923111 0.285820146559201
0.0493795275688171 0.277049802293986
0.0498912930488586 0.258908888527663
0.0501801371574402 0.25391821730193
0.0503083765506744 0.246038911456273
0.0505241453647614 0.237771565909968
0.0506033897399902 0.232799145751006
0.052994966506958 0.225824977195456
0.0567377507686615 0.23009845014644
0.0595249533653259 0.239130464664162
0.0632452070713043 0.2434529530536
0.0658179521560669 0.247808589201155
0.070850670337677 0.249179601547626
0.0749115347862244 0.249153752777174
0.0764316618442535 0.252171051598993
0.0776018500328064 0.248132626776737
0.0819774568080902 0.241916564146968
0.0838373005390167 0.237762635657316
0.0866940021514893 0.232797245761453
0.0885289907455444 0.22770656440062
0.0900700390338898 0.219857299042784
0.092154860496521 0.214608420379191
0.0935482978820801 0.209560881664817
0.0944430232048035 0.204194244487612
0.0958844125270844 0.201762618487009
0.096635639667511 0.200097093837549
0.098035991191864 0.198361220434826
0.0985373258590698 0.196804138372283
0.099325954914093 0.193898401678373
0.099760115146637 0.191651444523193
0.100031822919846 0.190697888694068
0.100451588630676 0.188624354537767
0.100765228271484 0.186543875167808
0.102451592683792 0.184549978864741
0.104859203100204 0.183825172863355
0.106842517852783 0.183559922389262
0.11184886097908 0.182772804466225
0.114052981138229 0.183030032812874
0.120038062334061 0.183222641599445
0.12308606505394 0.184179261292043
0.125674605369568 0.184573803551966
0.129112303256989 0.184514307086237
0.131533324718475 0.184454150461179
0.134348630905151 0.184332906638069
0.136461555957794 0.183709047169064
0.138635963201523 0.182589278663684
0.140150368213654 0.181808032287948
0.141961991786957 0.180993956064998
0.143644660711288 0.180400602399916
0.14465257525444 0.180174842913819
0.146185874938965 0.17974604487146
0.147138774394989 0.179599546451007
0.148100405931473 0.179191217664756
0.148520052433014 0.178632150731812
0.14928150177002 0.178461278728805
0.149894714355469 0.178462971060942
0.15011739730835 0.177889186011745
};
\end{axis}

\end{tikzpicture}
	\caption{\textbf{Depth Estimate Convergence}. 
		We plot the depth estimate corresponding to the Chips Can result in Figure~\ref{fig:front}.
		ClickBot actively estimates an object's depth as it approaches for grasping, and this depth estimate convergences as the camera moves closer and collects more data.
		Notably, the object's depth relative to the camera decreases with camera movement.
	}
	\label{fig:depth}
\end{figure}
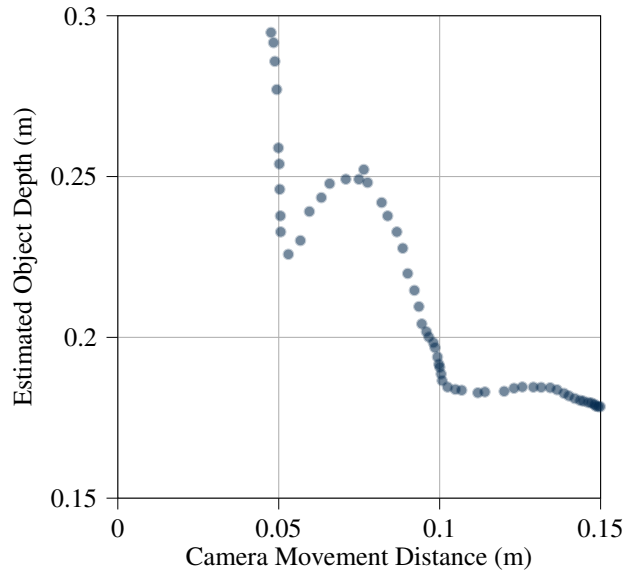

\begin{center}
	\Large\textbf{Supplementary Material:}\\
	\vspace{2mm}
	\Large\textbf{Mobile Robot Manipulation using Pure Object Detection}
\end{center}

\vspace{2mm}
\noindent \textbf{Per-Object TFOD Benchmark Results}.
We provide \textit{per-object} \textbf{T}ask-Focused \textbf{F}ew-Shot \textbf{O}bject \textbf{D}etection (TFOD) benchmark results in Figure~\ref{fig:TFOD_bar}, which correspond to the ClickBot $k=1,2,4$ few-shot example configurations in Table~\ref{tab:tfod}.
As in Table~\ref{tab:tfod}, we find opportunities for innovation across all settings, especially one- or two-shot detection.
The Wood, Chips Can, and Box of Sugar are particularly inaccurate for $k<4$.
Unsurprisingly, the $k=4$ configuration has the best performance for all objects with the exception of Gelatin.

We expect performance improvements across all objects and few-shot configurations with future few-shot object detection research.
In practice, such methodological advances will also improve robot task performance and reduce overall annotation requirements.

\vspace{2mm}
\noindent \textbf{Camera Movement and Learned Visual Servo Control}.
We plot the camera movements for learning visual servo control in Figure~\ref{fig:learn_vs_cam} with the corresponding learned parameters originally shown in Figure~\ref{fig:learn_param}.

For camera motion ($\Delta \mathbf{x}$), ClickBot repeats eight movement \textit{commands} comprising the permutations of \{-5, 0, 5\} \textrm{cm} across the $x$ and $y$ axes (e.g., $x =$ -5, $y =$ 5 for the second Broyden update).
However, ClickBot's base movements are imprecise for small motions, so the \textit{actual} measured movement distance we use for the update is slightly less (e.g., Base Forward $=$ -2.7~\textrm{cm} and Base Lateral $=$ 2.5~\textrm{cm} of actual motion for the second update).
Nonetheless, the actual motion profile is sufficient to learn ClickBot's visual control, which we use for all experiments in Section~\ref{sec:experiments}.

\vspace{2mm}
\noindent \textbf{Depth Estimate Convergence}.
In Section~\ref{sec:depth}, we introduce ClickBot's active depth estimation, which continually processes incoming data while approaching objects for grasping.
We provide an example depth convergence plot in Figure~\ref{fig:depth}, which corresponds to the Chips Can result in Figure~\ref{fig:front}.
ClickBot advances in 0.05~\textrm{m} increments, so the depth estimate generally completes with the object between 0.15~\textrm{m} to 0.2~\textrm{m} away.
In this example, after the grasp camera moves 0.15~\textrm{m}, the Chips Can's final estimated depth is 0.18~\textrm{m}, which leads to a successful grasp of the Chips Can.

As discussed in Section~\ref{sec:depth}, ClickBot estimates object depth from detection by comparing changes in bounding box size (i.e., optical expansion) with the corresponding camera movement, which we obtain using robot kinematics.
We use the Box$_{\text{LS}}$ equation \cite[(9)]{GrCo21} within our active depth estimation approach to process all available observations in a least-squares formulation, thus, our depth estimate generally improves as more data are collected.
Finally, the depth estimate's accuracy significantly improves as the object gets closer and exhibits more rapid optical expansion.

\vspace{2mm}
\noindent \textbf{Individual Trial Results for Task-Focused Annotation}.
We provide the task-focused few-shot annotation results for individual trials in Table~\ref{tab:learn_sup}.
All Mean results are the same as those originally shown in Table~\ref{tab:learn}. 
Remarkably, no experiment configuration uses more than a minute of human annotation time per object, which is approximately the same amount of time required to annotate a single segmentation mask and much less than the time required to generate a 3D object model.

We discuss a few notable individual trial results.
For the Visual Servo and Depth Benchmark on the Food: Chips Can, Potted Meat, Plastic Banana trial, ClickBot learns the Find, Move, and Depth tasks for all objects without prior annotation using 3 task-focused examples.
For Grasping \textit{in Clutter} with Prior Annotation on the Food: Box of Sugar, Tuna, Gelatin trial, ClickBot requires only 1 task-focused Move example to transfer learning from the prior grasp task to learn grasping in clutter.
Finally, for Grasping \textit{in Clutter} on the Food: Chips Can, Potted Meat, Plastic Banana trial, ClickBot learns all tasks for all objects in a cluttered setting without prior annotation using 7 task-focused examples.

\begin{figure*}
	\centering
	\includegraphics[width=0.99\textwidth]{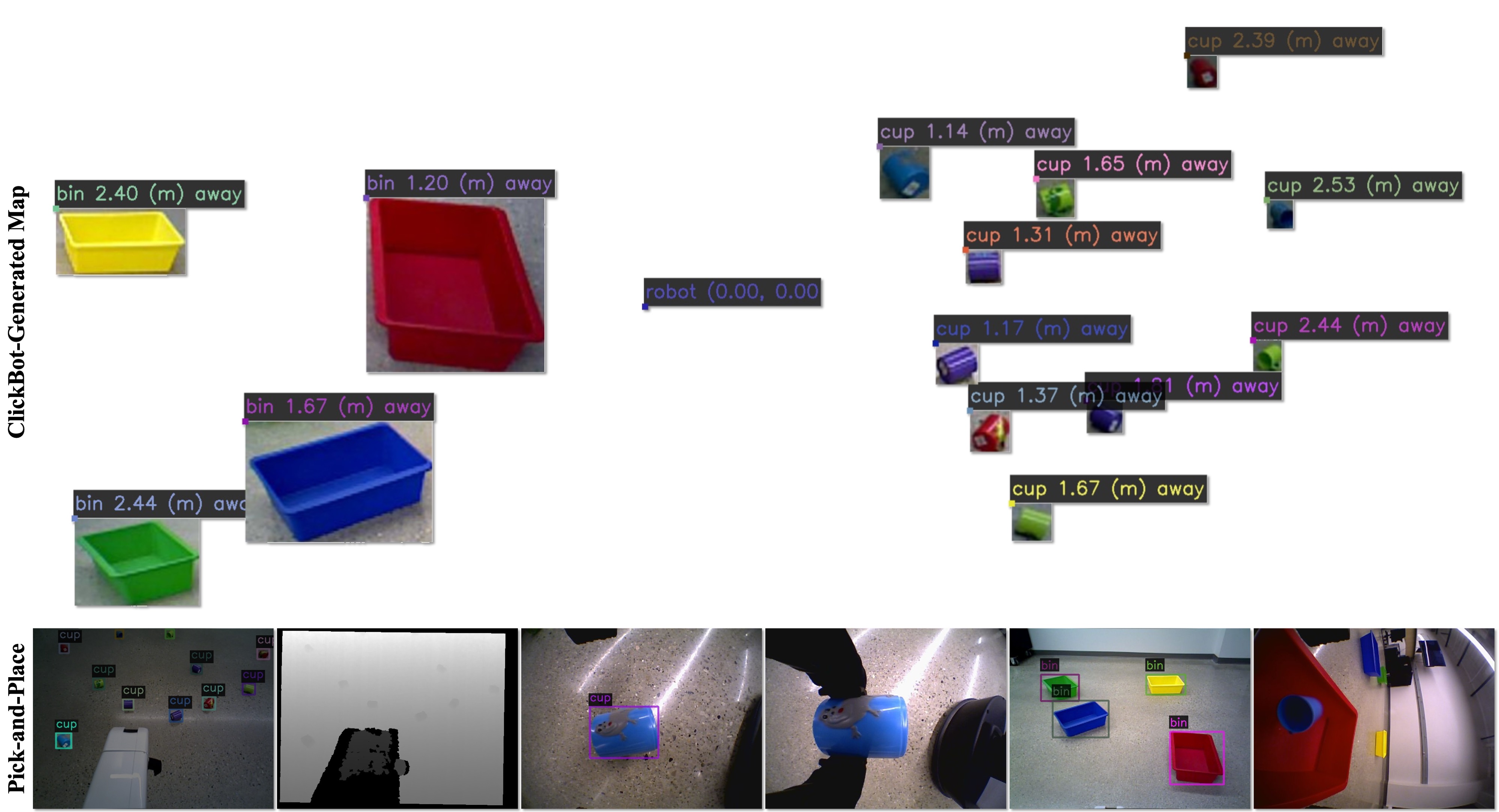}
	\caption{\textbf{ClickBot-Generated Map for Pick-and-Place with Dynamic Locations}. 
		In dynamic pick-and-place (bottom), ClickBot uses detection with an RGBD camera to locate and grasp scattered objects (left) and similarly uses detection to find a suitable placement location (right).
		Here, we show the ClickBot-generated map (top) corresponding to the pick-and-place result originally shown in Figure~\ref{fig:exp}.
	}
	\label{fig:rgbd}
\end{figure*}

\vspace{2mm}
\noindent \textbf{ClickBot-Generated Map for Dynamic Pick-and-Place}.
We provide an example ClickBot-generated map in Figure~\ref{fig:rgbd}, which corresponds to the dynamic pick-and-place result originally shown in Figure~\ref{fig:exp}.

ClickBot uses the same few-shot detection model with it's head-mounted RGBD camera, which enables ClickBot to map any RGB-based bounding box to a median 3D point using the corresponding depth image.
Using this map for the Find task, ClickBot quickly identifies the closest grasp object and subsequent placement location even after a grasped object is blocking ClickBot's grasp camera.

\begin{figure*}
	\centering
	\includegraphics[width=0.99\textwidth]{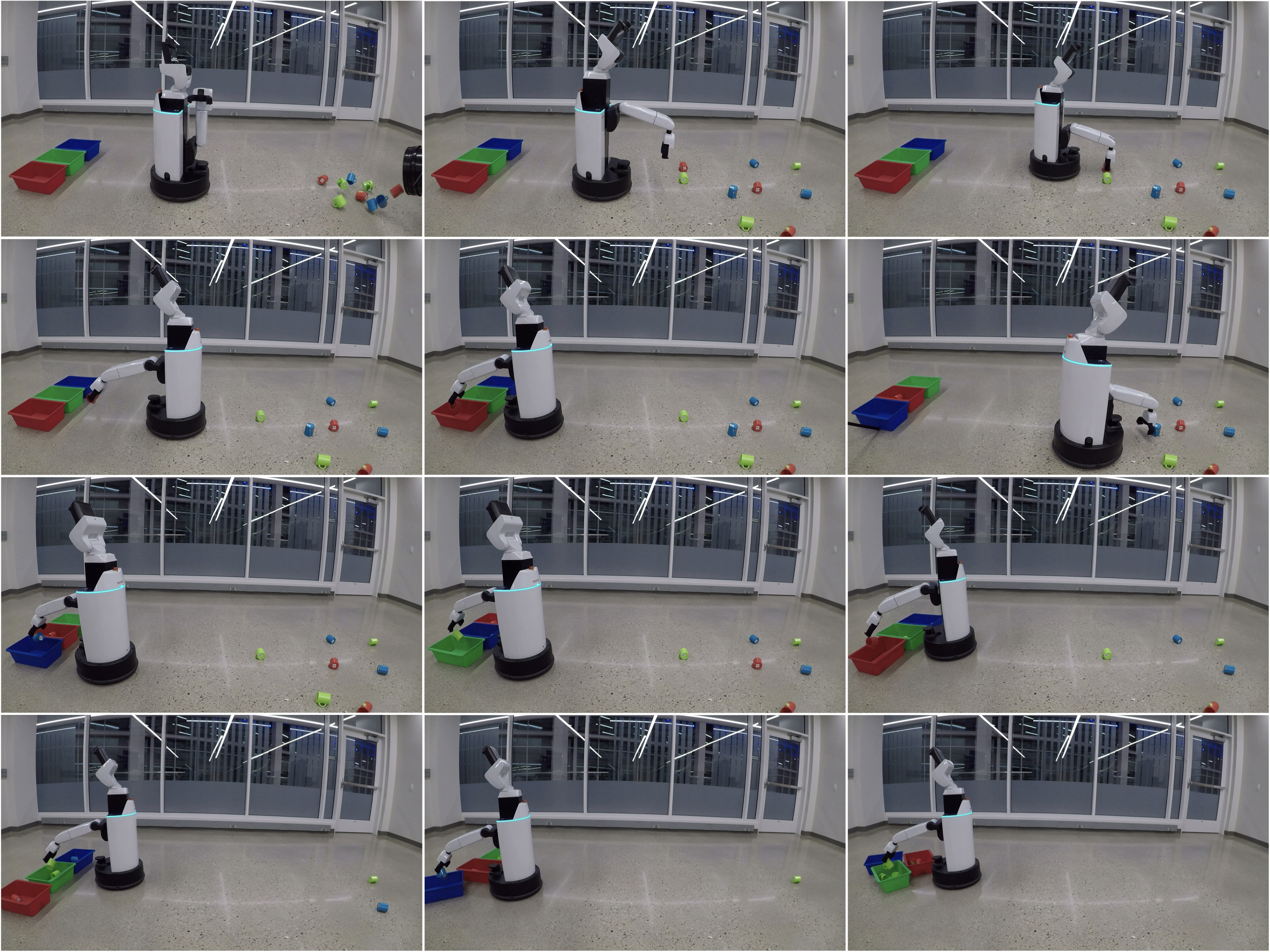}
	\caption{\textbf{Cleaning Scattered Cups with Dynamic Pick-and-Place}. 
		After we scatter cups on the ground, ClickBot uses its head camera to locate the cups and then its grasp camera to grasp the closest cup (first row, left to right).
		Next, ClickBot uses its head camera to locate the bins and then places the grasped red cup in the red bin to match color. 
		As ClickBot grasps the second cup, we also rearrange the bins (second row).
		ClickBot places each cup in the correctly colored bin, regardless of our rearranging the bins after each placement (third row).
		ClickBot cleans up all nine cups in 260~\textrm{seconds}, which equates to a rate of 124.6 mobile picks per hour (bottom row).
		To our knowledge, there is no precedent for this rate of vision-based mobile robot manipulation in the literature (see video at \url{https://youtu.be/giiSYDwZM4c}).
	}
	\label{fig:120}
\end{figure*}

\vspace{2mm}
\noindent \textbf{Cleaning Scattered Cups with Dynamic Pick-and-Place}.
We show ClickBot cleaning scattered cups with changing placement locations in Figure~\ref{fig:120}, which corresponds to the final dynamic pick-and-place experiment in Section~\ref{sec:exp_pph}.

\setlength{\tabcolsep}{8.3pt} 
\begin{table*}
	\centering
	\caption{
		\textbf{Task-Focused Few-Shot Annotation Results (Individual Trials)}. All results are from a single consecutive set of trials.
		Clicks are the number of annotated bounding boxes, which each require 7~\textrm{seconds} (see user study \cite{JaGr13}).
		Note that Clicks per Few-Shot Example varies with the number of in-view objects.
		CPU refers to training time. Mean results are the same as those originally shown in Table~\ref{tab:learn}. 
	}
	\footnotesize
	\begin{tabular}{| l | c | c | c | c | c | c | r | r | r |}
		\hline
		\cellcolor{rowblue} & \multicolumn{5}{c |}{\cellcolor{rowlearn} Number of Task-Focused} & \multicolumn{4}{c |}{\cellcolor{rowcost} Requirements Per Object Class} \\
		\hhline{| >{\arrayrulecolor{rowblue}}- >{\arrayrulecolor{black}}| >{\arrayrulecolor{rowlearn}} *{5}{-}>{\arrayrulecolor{black}}*{4}{|-}}
		\multicolumn{1}{| c |}{\cellcolor{rowblue}}& \multicolumn{5}{ c |}{\cellcolor{rowlearn} Few-Shot Examples Generated ($E$)} & \multicolumn{2}{c |}{\cellcolor{rowcost} Annotation}  & \multicolumn{1}{c |}{\cellcolor{rowcost} Robot}& \multicolumn{1}{c |}{\cellcolor{rowcost} CPU} \\
		\hhline{| >{\arrayrulecolor{rowblue}}- >{\arrayrulecolor{black}}| ---------}
		\multicolumn{1}{| c |}{\cellcolor{rowblue} Task-Focused Learning Experiment Trial} &  \cellcolor{rowlearn} Find & \cellcolor{rowlearn} Move &  \cellcolor{rowlearn} Depth & \cellcolor{rowlearn} Grasp & \cellcolor{rowlearn} Total & \multicolumn{1}{c|}{\cellcolor{rowcost} Clicks} &  \multicolumn{3}{c |}{\cellcolor{rowcost} Time (\textrm{seconds})}  \\
		\hline
		Learning Visual Control (Section~\ref{sec:exp_learn}) & 1 & 0 & N/A & N/A & 1 & 1.0 & 7.0   & 13.3 & 227  \\ 
		\hline \hline
		\multicolumn{10}{|c|}{\cellcolor{rowblue} Visual Servo and Depth Benchmark (Section~\ref{sec:exp_trial})} \\ \hline
		Tool: Power Drill, Marker, Padlock  & \bf 1	&	1	&	3	&	N/A		&	5	&	3.7	&	25.7	&	22.9	&	381  \\
		\rowcolor{rowgray} 	Tool: Wood, Spring Clamp, Screwdriver &\bf  1	&	1	&\bf 	2	&	N/A		&	4	&	3.7	&	25.7	&	10.9	&	309  \\
		Food: Chips Can, Potted Meat, Plastic Banana  &\bf  1	&\bf 	0	&\bf 	2	&	N/A		&\bf 	3	& \bf	2.7	& \bf	18.7	& \bf	9.3	&	 \bf233  \\
		\rowcolor{rowgray} Food: Box of Sugar, Tuna, Gelatin  &\bf  1	&	1	&	4	&	N/A		&	6	&	3.7	&	25.7	&	30.0	&	460  \\
		Kitchen: Mug, Softscrub, Skillet with Lid  &\bf  1	&	1	&	5	&	N/A	&	7	&	5.0	&	35.0	&	30.0	&	536  \\
		\rowcolor{rowgray} Kitchen: Plate, Spatula, Knife  &\bf  1	&\bf 	0	&	3	&	N/A	&	4	& \bf	2.7	& \bf	18.7	&	18.0	&	304 \\
		Shape: Baseball, Plastic Chain, Washer  & \bf 1	&	2	&	3	&	N/A		&	6	&	4.7	&	32.7	&	18.6	&	457 \\
		\rowcolor{rowgray} 	Shape: Stacking Cup, Dice, Foam Brick  &\bf  1	&	1	&	3	&	N/A		&	5	&	3.7	&	25.7	&	21.5	&	387  \\ \hline
		Mean
		& 1.0	&	0.9	&	3.1	&	N/A	&	5.0	 &	3.7	&	26.0	&	20.2	&	 383  	  \\ \hline \hline
		\multicolumn{10}{|c|}{\cellcolor{rowblue} Grasping with Prior Annotation (Section~\ref{sec:exp_grasp})} \\ 
		\hline
		Tool: Power Drill, Marker, Padlock  &  \bf 0	&	\bf 0	&\bf 	0	&	4	&	4	&	3.0	&	21.0	&	27.3	&	307 \\
		\rowcolor{rowgray} 	Tool: Wood, Spring Clamp, Screwdriver & \bf 0	&\bf 	0	&	1	&\bf 	2	&\bf 	3	&\bf 	2.7	&\bf 	18.7	&\bf 	21.9	&\bf 	231  \\
		Food: Chips Can, Potted Meat, Plastic Banana  & 1	&\bf 	0	&	1	&	3	&	5	&	3.7	&	25.7	&	32.3	&	378  \\
		\rowcolor{rowgray} Food: Box of Sugar, Tuna, Gelatin  & \bf 0	&	1	&	3	&\bf 	2	&	6	&	4.3	&	30.3	&	35.0	&	457  \\ \hline
		Mean
		& 0.3	&	0.3	&	1.3	&	2.8	&	4.5	&	3.4	&	23.9	&	29.1	&	343  	  \\ \hline 
		\hline
		\multicolumn{10}{|c|}{\cellcolor{rowblue} Grasping \textit{in Clutter} with Prior Annotation (Section~\ref{sec:exp_grasp})} \\ 
		\hline
		Tool: Power Drill, Marker, Padlock  &  1	&\bf 	0	&\bf 	0	&	3	&	4	&	2.7	&	18.7	&	32.6	&	374  \\
		\rowcolor{rowgray} 	Tool: Wood, Spring Clamp, Screwdriver &\bf  0	&	2	&\bf 	0	&	3	&	5	&	3.7	&	25.7	&	34.6	&	387 \\
		Food: Chips Can, Potted Meat, Plastic Banana  & 1	&\bf 	0	&	\bf 0	&	3	&	4	&	3.3	&	23.3	&	25.7	&	309  \\
		\rowcolor{rowgray} Food: Box of Sugar, Tuna, Gelatin  &\bf  0	&	1	&	\bf 0	&\bf 	0	&\bf 	1	&\bf 	1.0	&\bf 	7.0	&\bf 	0.2	&\bf 	76  \\ \hline
		Mean
		& 0.5	&	0.8	&	0.0	&	2.3	&	3.5	&	2.7	&	18.7	&	23.2	&	287  \\  \hline
		\hline
		\multicolumn{10}{|c|}{\cellcolor{rowblue} Grasping  (Section~\ref{sec:exp_grasp})} \\
		\hline
		Tool: Power Drill, Marker, Padlock  &  \bf 1	&	1	&	\bf 2	&	5	&	9	&	6.7	&	46.7	&	61.0	&	689  \\
		\rowcolor{rowgray} 	Tool: Wood, Spring Clamp, Screwdriver & \bf 1	&	1	&	\bf 2	&	\bf 3	&	7	&	6.0	&	42.0	&	38.3	&	543  \\
		Food: Chips Can, Potted Meat, Plastic Banana  & \bf 1	&\bf 	0	&	\bf 2	&\bf 	3	&	\bf 6	&\bf 	5.3	&\bf 	37.3	&	\bf 33.4	&	\bf 457  \\
		\rowcolor{rowgray} Food: Box of Sugar, Tuna, Gelatin  & \bf 1	&	1	&	4	&	4	&	10	&	6.0	&	42.0	&	73.0	&	770  \\ \hline
		Mean
		& 1.0	&	0.8	&	2.5	&	3.8	&	8.0	&	6.0	&	42.0	&	51.4	&	615   \\ \hline \hline
		\multicolumn{10}{|c|}{\cellcolor{rowblue} Grasping  \textit{in Clutter} (Section~\ref{sec:exp_grasp})} \\ \hline
		Tool: Power Drill, Marker, Padlock  & \bf 1	&	2	&	5	&	5	&	13	&	10.0	&	70.0	&	97.0	&	 1,008   \\
		\rowcolor{rowgray} 	Tool: Wood, Spring Clamp, Screwdriver & \bf 1	&\bf 	0	&	4	&	3	&	8	&\bf 	5.7	&\bf 	39.7	&	60.2	&	 614  \\
		Food: Chips Can, Potted Meat, Plastic Banana  & \bf 1	&	2	&	\bf 2	&\bf 	2	&\bf 	7	&	6.0	&	42.0	&\bf 	33.0	&\bf 	 540   \\
		\rowcolor{rowgray} Food: Box of Sugar, Tuna, Gelatin  & \bf 1	&	4	&	6	&	3	&	14	&	8.3	&	58.3	&	79.2	&	 1,082   \\ \hline
		Mean
		& 1.0	&	2.0	&	4.3	&	3.3	&	10.5	&	7.5	&	52.5	&	67.3	&	811   \\ \hline
	\end{tabular}
	\label{tab:learn_sup}
\end{table*}

\vspace{2mm}
\noindent \textbf{Application Novelty}.
We provide an extended discussion on application novelty to supplement the main paper.
We also provide an application comparison of related work in Table~\ref{tab:related}.
All citations are with the main paper's References.

A fundamental asset of robot perception is the opportunity to learn beyond static datasets from a robot's own surroundings.
Consequently, the robotics community has developed innovative solutions wherein robots actively learn in a fixed workspace.
To estimate an object's pose \cite{XiEtAl18}, robots interact with 3D-modeled objects \cite{YCB} to create new training data \cite{DeEtAl20,MaFlMaTe18}.
To grasp objects \cite{DuEtAl21}, robots perform large-scale data collection individually \cite{PiGu16} or with other robots \cite{LeEtAl18}.
To place objects, robots learn residual physics to toss objects into out-of-reach boxes \cite{ZeEtAl20}.
However, all of these robots learn in a fixed workspace, and scaling these solutions to mobile operation remains an open problem.
On the other hand, this paper uniquely addresses the problem of robot learning in a mobile application setting.
Mobile application challenges include moving cameras, changing environments, and dynamic grasp positioning for a mobile robot and dexterous workspace that move in the world frame.

For robot perception without learning, robot mobility has been achieved using closed-form visual servo control (VS), i.e., using visual data as input to a servo feedback control loop \cite{ChHu06,ChHu07,HuHaCo96}.
VS achievements include positioning UAVs \cite{GuHaMa08,McJaCo17} or wheeled robots \cite{LuOrGi08,MaOrPr07} and mobile manipulation \cite{KiEtAl16,WaLaSi10}.
However, all of these mobile VS robots use structured visual features (e.g., fiducial markers or LED panels).
On the other hand, this paper introduces a new detection-based approach to robot learning to learn VS, depth estimation, and grasping on a mobile robot in unstructured settings, thus extending VS to new applications where the environment and objects can change.

Achieving the mobile manipulation results in this paper required innovation across each of the detection-based tasks.
For detection-based visual servo control (Section~\ref{sec:boxvs}), we 1) develop a set of detection-based features that account for detection errors and multiple objects and 2) define a novel update formulation that learns visual servo control on average in less than 14~\textrm{s} and reduces learning variability by 65-85\% relative to prior visual servo approaches (Table~\ref{tab:vs_compare}).
For detection-based depth estimation (Section~\ref{sec:depth}), we adopt an existing least squares formulation \cite[(9)]{GrCo21} into a new active detection framework that improves depth estimation during the grasp approach while mitigating proximity-based detection errors. 
To our knowledge, this paper is the first work to use detection-based depth estimation in a real-time application.
For detection-based grasping (Section~\ref{sec:grasp}), we use a novel active multi-view grasp selection approach that requires only bounding boxes.
To our knowledge, this paper is the first work to grasp objects entirely from detection.
Finally, we introduce TFOD (Section~\ref{sec:tfod}) to learn all of these tasks for new objects and settings on a real robot in a variety of experiments (Section~\ref{sec:experiments}), thereby validating our detection-based method.

\vspace{2mm}
\noindent \textbf{Challenges, Solutions, and Future Work}.
We discuss a few implementation challenges we found in experiments and suggest some corresponding solutions and future work.

Moving HSR's grasp camera (Section~\ref{sec:exp_setup}) in close proximity to objects can cause the objects to blur, making detection more difficult.
However, if blur causes an error, ClickBot uses the blurred image for few-shot annotation to update its detection model, which we found improves detection performance on blurred images.
Another solution to decrease blur is to scale the control input $\mathbf{v}$ \eqref{eq:vs} to slow down the grasp camera when visual servoing to an object.

\setlength{\tabcolsep}{3pt} 
\begin{table} [t]
	\centering
	\caption{\textbf{Application Comparison of Related Work}.
		To our knowledge, this is the first work to use a real robot to learn few-shot mobile manipulation for novel objects.
	}
	\footnotesize
	\begin{tabular}{ l | c  c | c c | c | c | c |}
		\hhline{~ -------}
		& \multicolumn{4}{c |}{\cellcolor{rowlearn} Vision} & \multicolumn{3}{c |}{ \cellcolor{rowblue} Robot} \\
		\cline{2-8} 
		\parbox[t]{0mm}{\multirow{0}{*}{\rotatebox[origin=c]{90}{}}} &
		\parbox[t]{1mm}{\multirow{7}{*}{\rotatebox[origin=c]{90}{Few-Shot Object}}} & 
		\parbox[t]{2mm}{\multirow{7}{*}{\rotatebox[origin=c]{90}{Learning}}} &
		\parbox[t]{1mm}{\multirow{7}{*}{\rotatebox[origin=c]{90}{Non-Structured}}} & 
		\parbox[t]{2mm}{\multirow{7}{*}{\rotatebox[origin=c]{90}{Visual Features}}} & 
		\parbox[t]{2mm}{\multirow{7}{*}{\rotatebox[origin=c]{90}{Collect Train Data}}}& 
		\parbox[t]{2mm}{\multirow{7}{*}{\rotatebox[origin=c]{90}{Manipulate Objects}}} & 
		\parbox[t]{2mm}{\multirow{7}{*}{\rotatebox[origin=c]{90}{Mobile Operation}}}  \\
		\rule{0pt}{55pt} & & & & & & &   \\
		\hline
		\multicolumn{1}{| l | }{Few-Shot Object Detection e.g. \cite{ChEtAl18}} & \multicolumn{2}{c| }{{Yes}} &  \multicolumn{2}{c |}{{Yes}} 
		&  \multicolumn{1}{c| }{N/A} &  \multicolumn{1}{c| }{N/A} & \multicolumn{1}{c| }{N/A} \\  
		\rowcolor{rowgray} \multicolumn{1}{| l | }{Train Detector with Robot Data e.g. \cite{AlSuSu19}} & \multicolumn{2}{c| }{{\color{red} No}} & \multicolumn{2}{c| }{{Yes}}
		&  \multicolumn{1}{c| }{{Yes}} &  \multicolumn{1}{c| }{{\color{red} No}} & \multicolumn{1}{c| }{{Yes}} \\ 
		\multicolumn{1}{| l | }{Classic Visual Servo Control e.g. \cite{WaLaSi10}} & \multicolumn{2}{c| }{{\color{red} No}} & \multicolumn{2}{c|  }{{\color{red} No}}
		&  \multicolumn{1}{c| }{N/A} &  \multicolumn{1}{c| }{{Yes}} & \multicolumn{1}{c| }{{Yes}} \\  
		\rowcolor{rowgray} \multicolumn{1}{| l | }{Learned Visual Manipulation	e.g. \cite{JaEtAl19}} & \multicolumn{2}{c| }{{\color{red} No}} & \multicolumn{2}{c| }{{Yes}}
		&  \multicolumn{1}{c| }{{Yes}} &  \multicolumn{1}{c| }{{Yes}} & \multicolumn{1}{c| }{{\color{red} No}} \\  
		\multicolumn{1}{| l | }{Mobile Visual Manipulation e.g. \cite{GrFlCo20}} & \multicolumn{2}{c| }{{\color{red} No}} & \multicolumn{2}{c| }{{Yes}}
		&  \multicolumn{1}{c| }{{\color{red} No}} &  \multicolumn{1}{c| }{{Yes}} & \multicolumn{1}{c| }{{Yes}} \\  \hline
		\hline
		\rowcolor{rowgray} \multicolumn{1}{| l | }{ \bf ClickBot (Ours)} &  \multicolumn{2}{c| }{{\bf Yes}} &  \multicolumn{2}{c|}{{\bf Yes}} 
		&  \multicolumn{1}{c| }{{\bf Yes}} &  \multicolumn{1}{c| }{{\bf  Yes}} & \multicolumn{1}{c| }{{\bf  Yes}} \\
		\hline
	\end{tabular}
	\label{tab:related}
\end{table}

Our current approach to detection-based grasping (Section~\ref{sec:grasp}) uses an overhead antipodal grasp at the center of an object.
However, HSR's gripper (Section~\ref{sec:exp_setup}) is too small to grasp some of the YCB Dataset objects using this approach (e.g., the Skillet with Lid and Plate shown in Figure~\ref{fig:ycb}).
One solution is to train the detector to generate bounding boxes on \textit{only} the graspable portion of each object (e.g., the Skillet with Lid's handle).
Regardless, some objects are simply not graspable from overhead (e.g., the Plate).
In future work, we will expand our approach to include alternative grasp strategies (e.g., lateral grasping at an object's side) when the detected object is too large for overhead grasping.

Our baseline detection model is based on a Faster R-CNN \cite{ReEtAl15} configuration available on the popular and open source Detecton2 platform \cite{detectron2}.
As discussed in Section~\ref{sec:exp_setup}, one motivation for using this baseline was ease of reproducibility for our experimental results.
On the other hand, few-shot object detection (FSOD) is becoming a hotly studied area of object detection with increasingly rampant advances, even within just the past year 
\cite{Fan_2021_CVPR, Han_2021_ICCV, Hu_2021_CVPR, Li_2021_CVPR, Li2_2021_CVPR, Li3_2021_CVPR, Qiao_2021_ICCV, Sun_2021_CVPR, Wu_2021_ICCV, Zhang_2021_CVPR, Zhang_2021_WACV, Zhu_2021_CVPR}.
We openly admit that our results will likely improve with FSOD algorithms that are more advanced than our initial baseline approach.
While running more experiments with current and future FSOD algorithms to reduce annotation requirements and improve task performance is an area of future work, this paper currently provides a new TFOD Benchmark that makes robot-collected data and corresponding annotations publicly available for research.
Thus, with this paper, we are encouraging the object detection research community to join us in this effort to perform and evaluate methods in this new task-focused setting for robot manipulation, which will guide future innovation toward increasingly reliable few-shot detection for robotics applications.

\vspace{2mm}
\noindent \textbf{Supplementary Videos} are provided at \url{https://youtu.be/Bby4Unw7HrI}. 
Videos include a detection-based manipulation overview, the learning visual servo control experiment from Section~\ref{sec:exp_learn}, and two example dynamic pick-and-place experiments from Section~\ref{sec:exp_pph}, which includes ClickBot cleaning scattered objects with moving placement locations at over 120 picks-per-hour.

\newpage
\vspace{2mm}
\noindent \textbf{Acknowledgment}. Toyota Research Institute provided funds to support this work.

\end{document}